\documentclass[10pt,twocolumn,letterpaper]{article}

\PassOptionsToPackage{table}{xcolor}
\usepackage[pagenumbers]{iccv} 

\usepackage{bm}
\usepackage{booktabs}
\usepackage{multirow}
\usepackage{pifont}
\usepackage{tabularx}
\usepackage{rotating}
\usepackage{xcolor}
\usepackage{algorithm}
\usepackage{algpseudocode}
\usepackage{mathtools}
\usepackage{caption}
\usepackage[accsupp]{axessibility}

\definecolor{lightgray}{gray}{0.9}

\definecolor{iccvblue}{rgb}{0.21,0.49,0.74}
\usepackage[pagebackref,breaklinks,colorlinks,allcolors=iccvblue]{hyperref}

\def\paperID{13929} 
\def\confName{ICCV}
\def\confYear{2025}

\title{GVDepth: Zero-Shot Monocular Depth Estimation for Ground Vehicles based on Probabilistic Cue Fusion}

\author{
	Karlo Koledić \quad Luka Petrović \quad Ivan Marković \quad Ivan Petrović \\ [0.4cm]
	University of Zagreb Faculty of Electrical Engineering and Computing\\
	{\tt\small \{karlo.koledic, luka.petrovic, ivan.markovic, ivan.petrovic\}@fer.unizg.hr}
}

\begin{document}
\maketitle
\begin{abstract}
Generalizing metric monocular depth estimation presents a significant challenge due to its ill-posed nature, while the entanglement between camera parameters and depth amplifies issues further, hindering multi-dataset training and zero-shot accuracy.
This challenge is particularly evident in autonomous vehicles and mobile robotics, where data is collected with fixed camera setups, limiting the geometric diversity.
Yet, this context also presents an opportunity: the fixed relationship between the camera and the ground plane imposes additional perspective geometry constraints, enabling depth regression via vertical image positions of objects.
However, this cue is highly susceptible to overfitting, thus we propose a novel canonical representation that maintains consistency across varied camera setups, effectively disentangling depth from specific parameters and enhancing generalization across datasets.
We also propose a novel architecture that adaptively and probabilistically fuses depths estimated via object size and vertical image position cues.
A comprehensive evaluation demonstrates the effectiveness of the proposed approach on five autonomous driving datasets, achieving accurate metric depth estimation for varying resolutions, aspect ratios and camera setups.
Notably, we achieve comparable accuracy to existing zero-shot methods, despite training on a single dataset with a single-camera setup. Project website: \href{https://unizgfer-lamor.github.io/gvdepth/}{\textbf{https://unizgfer-lamor.github.io/gvdepth/}}

\end{abstract}    
\vspace{-10pt}
\section{Introduction}
\label{sec:intro}
Accurate 3D reconstruction is one of the fundamental computer vision challenges, with far-reaching applications across various fields.
Unlike multi-view approaches that rely on complex dense matching and pose estimation, learning-based Monocular Depth Estimation (MDE) exploits semantic and geometric cues to infer pixel-wise depth maps from a single image.
However, \emph{these cues are often specific to a narrow distribution of the training data}, resulting in poor generalization to out-of-distribution scenarios.
\begin{figure}
	\includegraphics[width=\linewidth, trim=35pt 10pt 25pt 10pt, clip]{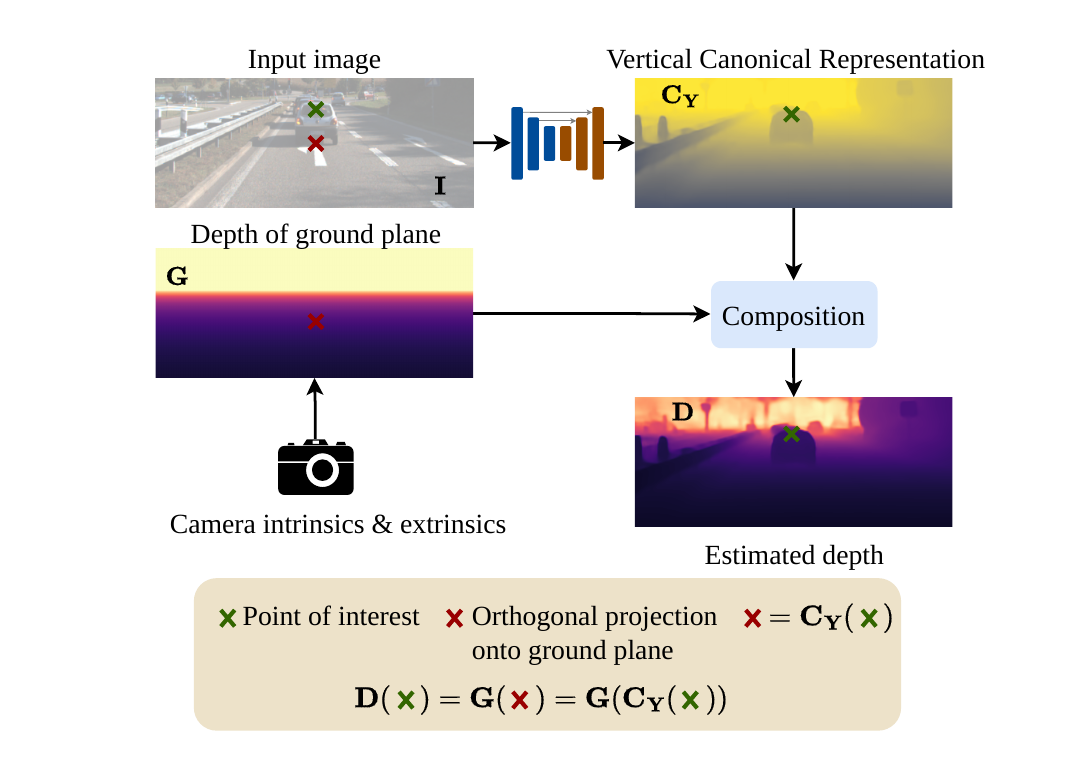}
	\caption{\textbf{Proposed Vertical Canonical Representation.} We introduce a novel intermediate representation that enables exploitation of underlying planar geometry priors, facilitating learning and generalization across varying camera setups. Under this representation, camera parameters and depth are completely disentangled, enabling depth regression via known depth of the ground plane.}
	\label{fig:intro}
	\vspace{-5pt}
\end{figure}

In addition to the environmental domain gap, MDE is particularly sensitive to the domain gap induced by the perspective geometry parameters of the camera system.
This issue is present in most supervised \cite{eigen2014depth, laina2016deeper, bhat2021adabins, yuan2022newcrfs, piccinelli2023idisc} and self-supervised methods \cite{zhou2017unsupervised, zhan2018unsupervised, godard2019digging,  guizilini20203d, zhou_diffnet, zhao2022monovit, wang2023planedepth} that are trained on a dataset collected with a single camera system.
Models tend to overfit to this perspective geometry bias, leading to significant performance degradation when inferring depth with different cameras.
Unfortunately, training a MDE model on data with a wide distribution of perspective geometries is far from trivial.
Due to the inherent ambiguity of the MDE problem and entanglement of depth and camera parameters \cite{piccinelli2024unidepth}, many methods resort to relative depth estimation with unknown shift and scale \cite{chen2016single, xian2018monocular, ranftl2020towards, yin2020diversedepth, yin2021learning, ke2024repurposing, yang2024depth}.
Recently, zero-shot metric depth estimation has been achieved with various approaches, including embedding \cite{guizilini2023towards}, estimating \cite{bochkovskii2024depth, piccinelli2024unidepth} or disentangling the camera parameters \cite{yin2023metric3d}.
While these methods present impressive results, \emph{they require large-scale training on data acquired from multiple camera systems}, which is not always practical or cost-effective.

One specific area with numerous applications across varying fields is MDE for ground-based vehicles and robots.
In this context, an additional perspective geometry constraint arises due to the fixed position of the camera relative to the ground plane.
This constraint induces an additional depth cue: the vertical image position of the ground-contact point, which has been shown to be a dominant cue in autonomous driving scenarios \cite{dijk2019neural}.

These findings motivate us to utilize the vertical image position cue and the inherent planarity of the scene in order to enable accurate depth estimation and improve generalization for arbitrary intrinsic and extrinsic camera parameters.
Contrary to the object size cue, widely used by the aforementioned zero-shot methods \cite{guizilini2023towards, bochkovskii2024depth, piccinelli2024unidepth, yin2023metric3d}, \emph{we argue that the vertical image position presents a cue that is more fundamentally rooted in the perspective geometry configuration of the scene}.
However, without proper modeling, this cue is highly susceptible to overfitting to the specific camera system present in the training data \cite{koledic2023gendepth, peng2021excavating, koledic2023moft}.

To leverage this cue, while enabling training with diverse perspective geometries and zero-shot generalization, in this paper we propose a method to ensure its consistency for arbitrary camera setups.
Specifically, we introduce a novel intermediate depth representation termed \textit{Vertical Canonical Representation}. 
Rather than directly regressing depth values, our model predicts vertical image positions of the orthogonal projections of points onto the ground plane, as visualized in \cref{fig:intro}.
By utilizing the known depth of the ground plane, this approach allows the mapping between the canonical representation and depth space to adapt with varying camera parameters, effectively disentangling the vertical position cue from specific perspective geometry.
To further address potential errors in certain regions, we introduce an architecture that adaptively fuses depth predictions from both vertical position and object size cues, guided by dynamically estimated uncertainties.

We comprehensively evaluate our method's zero-shot capability on five autonomous driving datasets, analyzing its robustness and accuracy across various cues. We demonstrate that our approach matches the accuracy of existing zero-shot methods, despite using a single datasets collected with a single camera system.

To summarize, we present the following contributions:
\begin{itemize}
	\item a novel canonical transformation that enables the exploitation of vertical image position cue across diverse perspective geometries, enhancing both training scalabilty and zero-shot generalization,
	\item a model architecture for the probabilistic fusion of depth maps estimated using object size and vertical image position cues,
	\item a thorough ablation of different depth cues, with insights previously unseen in the literature.
\end{itemize}

\section{Related Work}
\label{sec:rel_work}
\textbf{Domain-dependent Monocular Depth Estimation.} In a seminal work \cite{eigen2014depth}, learning-based MDE has been established as a supervised dense regression task via ground-truth data obtained by reprojection of LiDAR scans.
Subsequent works improve upon the original approach by introducing various architectural novelties, such as convolutional models with skip connections between encoder and decoder \cite{laina2016deeper}, integration of conditional random fields \cite{li2015depth, yuan2022newcrfs}, adversarial training \cite{jung2017depth, lore2018generative} or attention-based architectures \cite{piccinelli2023idisc, bhat2021adabins}.
Unfortunately, these methods are specifically fine-tuned for accuracy on well established indoor \cite{silberman2012indoor} and outdoor \cite{geiger2013vision} benchmarks, \emph{experiencing significant accuracy degradation under distribution shifts} \cite{koledic2023gendepth}. 

\noindent{}\textbf{Domain-agnostic Monocular Depth Estimation.} The inherent ambiguity and interdependence between camera parameters and depth present challenges for training a robust MDE model that generalizes across a wide range of environments.
As datasets are usually collected with a single camera system, models tend to act discriminatively and associate perspective geometry parameters with the pertaining environment.
Relative depth estimation methods \cite{chen2016single, xian2018monocular, ranftl2020towards, yin2020diversedepth, yin2021learning, yang2024depth, ke2024repurposing} bypass camera parameter dependency by estimating depth with an unknown scale. 
While these approaches achieve visually impressive results, \emph{their inability to estimate metric depth limits their applicability}, often necessitating integration with additional sensors.

ZoeDepth~\cite{bhat2023zoedepth} is the first work to show satisfactory metric accuracy across different domains by fine-tuning a pre-trained relative depth estimation model on a single dataset with metric scale.
ZeroDepth~\cite{guizilini2023towards} chooses a different approach by providing the model with embedding of camera parameters, effectively disambiguating the features from the specifics of the camera system.
Metric3D~\cite{yin2023metric3d} addresses the focal length-depth ambiguity by estimating depth in a canonical space defined through focal length denormalization.
UniDepth \cite{piccinelli2024unidepth} introduces a universal approach, simultaneously predicting both the depth and camera parameters, enabling metric depth estimation for images acquired with unknown sensors.
Although these methods demonstrate impressive results, \emph{they rely on abundance of training data and computational resources}.
Furthermore, they are not truly zero-shot, as they tend to overfit to the typically high training resolution, hindering real-time performance on resource-constrained devices that require low-resolution estimation.

\noindent{}\textbf{Planar Constraints for Depth Estimation.} Real-world scenes are often highly regular, featuring numerous planar structures that can serve as additional cues to guide depth estimation \cite{shao2023nddepth, patil2022p3depth}.
A highly relevant application of MDE is in autonomous driving and mobile robotics, where a fixed relation between the camera system and the ground plane induces additional perspective geometry constraints.
It has been shown that models heavily utilize the corresponding ground-contact cues, completely disregarding the object size cue \cite{dijk2019neural}.
Moreover, incorporating ground plane constraint within the model enhances overall model accuracy, both in supervised \cite{yang2023gedepth}, and self-supervised training \cite{cecille2024groco, wang2023planedepth}. However, these methods leverage the known perspective geometry constraints to enhance depth estimation only for road pixels, leading to limited generalization capabilities.

\section{Proposed Method}
\label{sec:method}
\textbf{Preliminaries.} In this work, we examine an extremely common application domain of MDE: ground vehicles or a robot with a fixed camera configuration.
Assuming an ego coordinate system $\bm{\mathcal{F}}_\mathcal{E}$ $(x\rightarrow,y{\scriptstyle\nearrow},z\odot)$ defined at the ground level, and camera coordinate system $\bm{\mathcal{F}}_\mathcal{C}$ $(x\rightarrow,y{\;\otimes},z{\scriptstyle\nearrow})$, we consider a camera system parameterized by intrinsic matrix $\mathbf{K}$ and extrinsic matrix $^{\mathcal{C}}_{\mathcal{E}}[\mathbf{R|t}]$, where:
\begin{equation}
\begingroup 
\setlength\arraycolsep{4pt}
\mathbf{K} = \begin{bmatrix}
	f_x & 0 & c_x \\
	0 & f_y & c_y \\
	0 & 0 & 1
\end{bmatrix},
\endgroup
\quad ^{\mathcal{C}}_{\mathcal{E}}\mathbf{R} =
\begingroup 
\setlength\arraycolsep{4pt}
\begin{bmatrix}
1 & 0 & 0\\
0 & \sin \theta & -\cos \theta\\
0 & \cos \theta & \sin \theta\\
\end{bmatrix},
\endgroup
\label{eqn:params}
\end{equation}
and $\mathbf{t} = [0, 0, -h]^T$,
with $(\theta, h)$ representing camera pitch and height, respectively.
Transformation of points from ego to camera coordinate frame is then: $\mathbf{p}_{\mathcal{C}} = \prescript {\mathcal{C}}{\mathcal{E}}{\mathbf{R}}\mathbf{p}_{\mathcal{E}} + \mathbf{t}$.
We disregard the camera yaw and roll in our parameterization; yaw has no effect on our canonical transformations, while roll is typically negligible in real-world setups.

\noindent{}\textbf{Problem Statement.}
We aim to learn a dense mapping $f:(\mathbf{I}, \bm{\mathcal{C}}) \mapsto \mathbf{D}$, with $\mathbf{I} \in \mathbb{R}^{H \times W \times 3}, \mathbf{D} \in \mathbb{R}^{H \times W}$ being RGB image and a correspondending depth map, and $\bm{\mathcal{C}}$ representing camera parameterization in \cref{eqn:params}.
Most importantly, the learned mapping should be able to generalize to images acquired with arbitrary camera system.
In contrast to existing zero-shot methods \cite{piccinelli2024unidepth, guizilini2023towards, bochkovskii2024depth, yin2023metric3d}, which rely on multiple datasets with varying camera parameters, we aim to learn this mapping using data acquired with a single camera system $\bm{\mathcal{C}}_{\text{Train}}$.

\subsection{Depth Cues}
\begin{figure}[!t]
	\includegraphics[width=\linewidth]{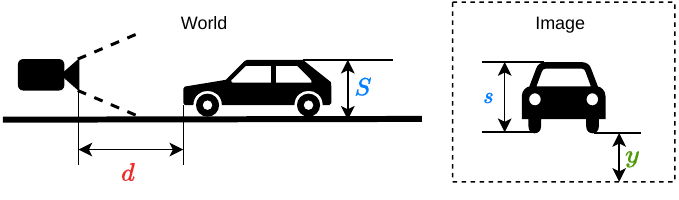}
	\caption{\textbf{Depth cues in MDE for ground vehicles}. Depth $d$ can be calculated both as a function of imaging size $s$ (with corresponding known object size $S$) and as a function of vertical image position $y$.}
	\label{fig:cues}
	\vspace{-8pt}
\end{figure}
Monocular depth estimation relies on semantic and geometric cues in the training data to solve the depth ambiguity of monocular perspective geometry.
As illustrated in \cref{fig:cues}, one of the possible cues to regress depth $d$ is the relation between object's real-world size $S$ and imaging size $s$:
\begin{equation}
	d=f_y \frac{S}{s}.
\end{equation}
However, this cue leads to entanglement between focal length and depth, leading to generalization issues when camera parameters change.
Recent zero-shot methods address this issue by embedding \cite{guizilini2023towards}, estimating \cite{bochkovskii2024depth, piccinelli2024unidepth} or disentangling the focal length \cite{yin2023metric3d}.
A major drawback of this cue is its reliance on the object's real world size $S$, which is highly dependent on the particular environmental domain. 

For ground vehicles, a fixed position of the camera system relative to the ground introduces an additional perspective geometry constraint -- known depth of the ground plane, which is constant for all images acquired by the same camera setup.
For each pixel $(u, v)$, depth $d$ is calculated from the intersection of the camera rays $d \,\mathbf{K}^{-1}[u, v, 1]^T$ and the ground plane $(^{\mathcal{C}}_{\mathcal{E}}\mathbf{R}\mathbf{n}_{\mathcal{E}})^T\mathbf{p}_{\mathcal{C}} +h = 0$, with $\mathbf{n}_{\mathcal{E}} = [0, 0, 1]^T$ being the upward ground normal.
This constraint enables depth regression via object's ground contact point:

\begin{equation}
	\begin{aligned}
		d &= \frac{-h}{(^{\mathcal{C}}_{\mathcal{E}}\mathbf{R}\mathbf{n}_{\mathcal{E}})^T\mathbf{K}^{-1}[u, v, 1]^T} \\
		&= \frac{f_y h}{\left(H-c_y-y\right) \cos (\theta)-f_y \sin (\theta)},
		\label{eqn:vertical}
	\end{aligned}
\end{equation}
where $y = H - v$ represents vertical image position of the ground projection, as visualized in \cref{fig:cues}.
Compared to the object size cue, which is entirely semantic, this cue implicitly encodes the known perspective geometry constraints of the scene, thereby inducing a valuable prior to the ill-posed MDE problem.
\emph{However, it is highly reliant on accurate and consistent camera calibration.}

\subsection{Canonical Transforms}
\begin{figure*}
	\includegraphics[width=0.9\linewidth]{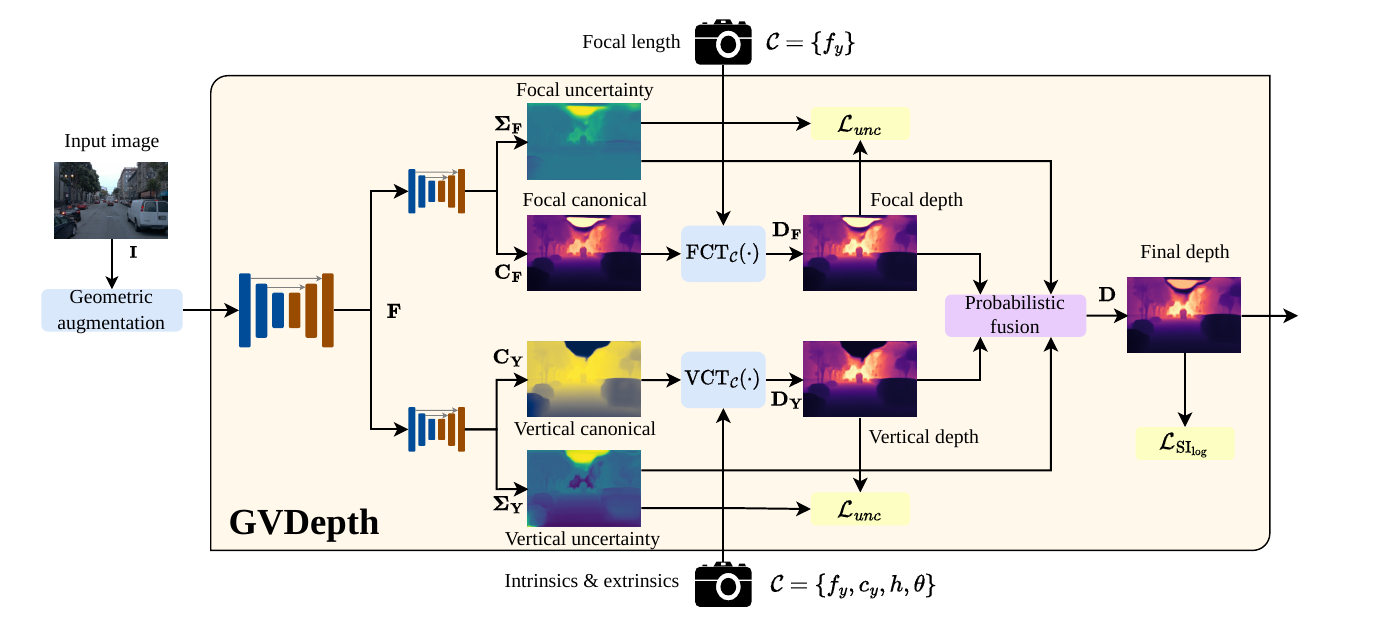}
	\centering
	\caption{\textbf{Model architecture}. Given the input image and corresponding camera parameters, our model predicts two canonical depth representations, $(\mathbf{C_F}, \mathbf{C_Y})$ and corresponding uncertanties $(\mathbf{\Sigma_F}, \mathbf{\Sigma_Y})$. Canonical representations are disentangled from the corresponding camera parameters, enabling separate exploitation of object size cue and vertical image position cue. Canonical representations are then transformed into depth maps $(\mathbf{D_F}, \mathbf{D_Y})$, which are subsequently adaptively fused into a final depth map $\mathbf{D}$ based on estimated uncertanties.}
	\label{fig:model}
	\vspace{-10pt}
\end{figure*}
Metric3D \cite{yin2023metric3d} proposed a canonical transform disentangling the predicted depth and camera parameters, enabling the use of object size cue for arbitrary camera systems.
To achieve this, depth is predicted in canonical space $\mathbf{C_F}$ with a canonical focal length $f_c$, and post-processed to metric depth $\mathbf{D_F}$ by applying Focal Canonical Transform $\mathrm{FCT}_{\bm{\mathcal{C}}}(\cdot)$, where $\mathbf{D_F} =\mathrm{FCT}_{\bm{\mathcal{C}}}(\mathbf{C_F}) =\frac{f_y}{f_c}\mathbf{C_F}$, and $\bm{\mathcal{C}} = \{f_y\}$.

Besides the object size cue, in this paper we propose to leverage the vertical image position cue, effectively encoding the configuration of the scene's perspective geometry.
Unfortunately, this cue also induces entanglement between camera parameters and depth.
To enhance cross-dataset generalization, we introduce an additional canonical transform, which enables consistent exploitation of the ground plane prior.
Under this formulation, the model estimates a dense map $\mathbf{C_Y}$ representing vertical image position of the ground projection point.
Metric depth $\mathbf{D_Y}$ is then obtained via Vertical Canonical Transform $\mathrm{VCT}_{\bm{\mathcal{C}}}(\cdot)$, where $\mathrm{VCT}_{\bm{\mathcal{C}}}(\mathbf{C_Y})$ corresponds to mapping $y \mapsto d$ obtainable from \cref{eqn:vertical}.
Therefore, this transformation requires knowledge of camera parameters $\bm{\mathcal{C}} = \{f_y, c_y, h, \theta\}$.
Crucially, \emph{the proposed canonical space remains consistent under variations of camera parameters, facilitating generalization to diverse scenarios.}

The proposed $\mathrm{VCT}_{\bm{\mathcal{C}}}(\cdot)$ is related to the depth of the ground plane, frequently used to enhance MDE accuracy in autonomous driving scenarios \cite{yang2023gedepth, cecille2024groco, wang2023planedepth}.
However, these methods exploit the ground-plane constraint exclusively for road pixels, resulting in limited generalization capability.

\subsection{Model Architecture}

Both cues have their advantages and drawbacks. For example, vertical position cue is highly consistent across environments due to the reduced reliance on semantic information.
However, it is highly sensitive to calibration errors, and performs worse as depth increases. An ideal model should leverage both cues, taking into account the confidence for respective image regions.
To that end, as visualized in \cref{fig:model}, we design our model to predict depth maps with both cues separately, and then fuse them into a final depth map based on the estimated aleatoric uncertainty.

We use a standard encoder-decoder model with skip connections to process and upsample image features, gaining $\mathbf{F} \in \mathbb{R}^{h \times w \times C}$, where $(h,w) = (\frac{H}{2}, \frac{W}{2})$.
We forward these features to two very small UNet networks \cite{ronneberger2015u}, producing our canonical representations $(\mathbf{C_F}, \mathbf{C_Y})$  and accompanying uncertainty estimates $(\mathbf{\Sigma_F}, \mathbf{\Sigma_Y})$.
Canonical representations are forwarded to $\text{FCT}(\cdot)$ and $\text{VCT}(\cdot)$, obtaining $(\mathbf{D_F}, \mathbf{D_Y})$, which are then fused into a final depth map $\mathbf{D} \in \mathbb{R}^{h \times w}$:
\begin{equation}
\mathbf{D} = \frac{\mathbf{\Sigma_Y} \odot \mathbf{D_F} + \mathbf{\Sigma_F} \odot \mathbf{D_Y}}{\mathbf{\Sigma_Y} + \mathbf{\Sigma_F}}.	
\label{eqn:fusion}
\end{equation}

\subsection{Geometric Augmentation}
Contrary to zero-shot methods trained on multiple datasets \cite{piccinelli2024unidepth, guizilini2023towards, bochkovskii2024depth, yin2023metric3d}, we aim to achieve similar generalization capabilities by training on a single camera system.
However, without geometric diversity induced by varying camera parameters, model can simply disregard our proposed canonical transformations.
Therefore, we apply geometric augmentations, namely random cropping and resizing, to simulate different camera perspectives.
Crucially, compared to \cite{piccinelli2024unidepth, bochkovskii2024depth, yin2023metric3d}, we train our model with multiple resolutions, enabling generalization for arbitrary image sizes.

\subsection{Optimization}
Our main training objective is to guide the fused depth map $\mathbf{D}$ towards the ground-truth depth $\mathbf{D}^*$. Given the distance in the $\mathrm{log}$ domain $\mathbf{E} = \log(\mathbf{D}) - \log(\mathbf{D}^*)$, the $\mathrm{SI}_{\mathrm{log}}$ loss \cite{eigen2014depth} is defined as:

\begin{equation}
	\mathcal{L}_{\mathrm{SI}_{\log }}=\alpha \sqrt{\mathbb{V}[\mathbf{E}]+\lambda \mathbb{E}^2[\mathbf{E}]},
\end{equation}
with $\alpha = 10$ and $\lambda = 0.15$ as in \cite{bhat2021adabins}, and $\mathbb{V}$, $\mathbb{E}$ being variance and expectation, respectively.

Additionally, we enhance depth estimation accuracy from both canonical transformations by employing an uncertainty-weighted loss:
\begin{align}
	\mathcal{L}_{\text{unc}} = & \, \mathbf{\Sigma_F}^{-1} \odot ||\mathbf{D_F - D^*}||_1 + \mathbf{\Sigma_Y}^{-1} \odot ||\mathbf{D_Y - D^*}||_1 \nonumber \\
	& + \log(\mathbf{\Sigma_F}) + \log(\mathbf{\Sigma_Y}),
	\label{eqn:unc}
\end{align}
which is equivalent to maximizing the log-likelihood, but with L1 norm instead of L2.
This approach down-weights regions where each cue exhibits higher uncertainty, encouraging each cue to focus on specific parts of the image where it performs better.
To facilitate training stability \cite{kendall2017uncertainties}, we predict $\log(\cdot)$ uncertainties, and perform $\exp(\cdot)$ operation accordingly.

Finally, same as in \cite{piccinelli2024unidepth}, we employ geometric consistency loss. Given two different sets of geometric augmentations of the same image ($\mathcal{T}_1$, $\mathcal{T}_2$), geometric consistency loss is defined as: 
\begin{equation}
\mathcal{L}_{\text{con}} = \left\|\mathcal{T}_2 \circ \mathcal{T}_1^{-1} \circ\mathbf{D}_1 -\operatorname{sg}\left(\mathbf{D}_2 \right)\right\|_1,
\end{equation}
where $\mathbf{D}_i$ are estimated depth maps for i'th augmentation, and $\operatorname{sg}(\cdot)$ corresponds to the stop-gradient operation.

Our final loss is a linear combination:
\begin{equation}
	\mathcal{L} = \mathcal{L}_{\mathrm{SI}_{\log }} + \lambda_{\text{unc}}\mathcal{L}_{\text{unc}} + \lambda_{\text{con}}\mathcal{L}_{\text{con}}, 
\end{equation}
with $\lambda_{\text{unc}} = 0.5$ and $\lambda_{\text{con}} = 0.1$.

\section{Experiments}
\label{sec:experiments}
\noindent{}\textbf{Datasets.}
We use a diverse set of autonomous driving datasets, each collected with a different camera setup.
Specifically, we utilize an aggregation of KITTI \cite{geiger2013vision}, DDAD \cite{guizilini20203d}, DrivingStereo \cite{yang2019drivingstereo}, Waymo \cite{Sun_2020_CVPR} and Argoverse Stereo \cite{Argoverse} datasets.
Due to the missing and sometimes inaccurate information, we calibrate extrinsic parameters, i.e., camera height and pitch, which are required for our canonical representation.
Further details are provided in the supplementary materials.

\noindent{}\textbf{Implementation Details.}
We use a standard encoder-decoder with skip connections \cite{godard2019digging}. 
In all experiments we use a ConvNext-Large \cite{liu2022convnet} encoder initialized with ImageNet weights.
To provide the model with additional contextual information, we fuse ground plane embedding in the decoder layers \cite{koledic2023gendepth, yang2023gedepth}.
In our depth cue fusion model, secondary U-Nets are designed with a lightweight structure, containing only three convolutional layers for both downsampling and upsampling.

Our models are trained in PyTorch \cite{paszke2019pytorch} with an AdamW optimizer \cite{DBLP:conf/iclr/LoshchilovH19}.
Learning rate is progressively reduced with Cosine Annealing from 1e-4 to 1e-5.
For all datasets, we use random cropping and resizing to $H \in [200, 500]$, with $W$ chosen to preserve the aspect ratio after cropping.
To increase resolution diversity in a single batch update, we perform gradient accumulation for 2 training steps.
All models are trained for 150000 optimizer steps with an effective batch size of 16, on a single NVIDIA RTX A6000 GPU.
Compared to concurrent zero-shot methods \cite{piccinelli2024unidepth, guizilini2023towards, yin2023metric3d}, this constitutes a fairly simple training setup, both in computational and data resources, demonstrating the efficiency of our approach.

\noindent{}\textbf{Evaluation Details.}
We use established depth metrics: absolute relative error ($\mathrm{A.Rel}$), root mean squared error ($\mathrm{RMS}$), root mean squared log error ($\mathrm{RMS}_{\log})$ and accuracy $\mathrm{\delta}_i$ under a threshold $1.25^i$. Depth is evaluated in the metric range of [0\,m, 80\,m].

We evaluate GVDepth at an image resolution of $H \times 640$, where $H$ varies according to the dataset to preserve the original aspect ratio. For a fair comparison, domain-specific models are trained and tested using the same image dimensions, which depend on specific dataset.
All models are evaluated using a ConvNext-L backbone if available; otherwise, we select the backbone with the most similar complexity.
For zero-shot methods, we use the provided checkpoints.
For metric zero-shot models which are optimized for a specific training resolution, we report results at both their original training resolution and our testing resolution for a comprehensive evaluation.

\subsection{Evaluation of Zero-shot Accuracy}
\begin{table*}[t]
	\centering
	\caption{\textbf{Generalization capability comparison.} Domain-dependent methods \cite{godard2019digging, zhou_diffnet, yuan2022newcrfs, piccinelli2023idisc} are trained and evaluated on our evaluation setup at an $H \times 640$ resolution. Zero-shot methods \cite{guizilini2023towards, piccinelli2024unidepth, yin2021learning, yin2023metric3d, ranftl2020towards} are evaluated with provided checkpoints. (\dag): Shift and scale alignment with ground-truth. (\ddag): Resizing and padding to native training resolution. Best zero-shot results without resolution alignment are \textbf{bolded}, second best are \underline{underlined}. In-domain evaluation results are \colorbox{lightgray}{shaded}.}
	\vspace{-5pt}
	\resizebox{\linewidth}{!}{
		\begin{tabular}{l|c|c|cccccc|cccccc}
			\toprule[0.4mm]
			\multirow{2}{*}{\textbf{Method}} & \textbf{Train} & \textbf{\#~Train}& \multicolumn{6}{c|}{\textbf{KITTI}} & \multicolumn{6}{c}{\textbf{DDAD}} \\
			& \textbf{dataset} & \textbf{samples} &  $\mathrm{A.Rel}\downarrow$ & $\mathrm{RMS}\downarrow$ & $\mathrm{RMS}_{\log}\downarrow$ &  $\mathrm{\delta}_{1}\uparrow$ & $\mathrm{\delta}_{2}\uparrow$ & $\mathrm{\delta}_{3}\uparrow$ & $\mathrm{A.Rel}\downarrow$ & $\mathrm{RMS}\downarrow$ & $\mathrm{RMS}_{\log}\downarrow$ &  $\mathrm{\delta}_{1}\uparrow$ & $\mathrm{\delta}_{2}\uparrow$ & $\mathrm{\delta}_{3}\uparrow$\\
			\midrule
			
			Monodepth2~ \cite{godard2019digging} &  \multirow{4}{*}{KITTI}& \multirow{4}{*}{23K}& \cellcolor{lightgray}9.02&\cellcolor{lightgray}3.94&\cellcolor{lightgray}0.137& \cellcolor{lightgray}91.4&\cellcolor{lightgray}98.3&\cellcolor{lightgray}99.5 & 37.8& 14.6& 0.599& 20.1& 45.7& 74.0\\
			DIFFNet ~\cite{zhou_diffnet}& & & \cellcolor{lightgray}8.15&\cellcolor{lightgray}3.86&\cellcolor{lightgray}0.129&\cellcolor{lightgray}92.0&\cellcolor{lightgray}98.4&\cellcolor{lightgray}99.6& 37.3& 14.4& 0.563& 16.1& 42.6& 79.1 \\
			NeWCRFs ~\cite{yuan2022newcrfs} & & &\cellcolor{lightgray}5.51&\cellcolor{lightgray}2.20&\cellcolor{lightgray}0.084&\cellcolor{lightgray}97.3&\cellcolor{lightgray}99.7&\cellcolor{lightgray}99.9 &33.7 & 15.6& 0.571& 30.2& 62.9  & 79.2\\
			iDisc~ \cite{piccinelli2023idisc} & & &\cellcolor{lightgray}5.33&\cellcolor{lightgray}2.11&\cellcolor{lightgray}0.081&\cellcolor{lightgray}97.1 &\cellcolor{lightgray}99.7&\cellcolor{lightgray}99.9&27.4& 13.3& 0.461& 38.4& 74.3& 88.4 \\
			
			\midrule
			MiDaS\textsuperscript{\dag}~\cite{ranftl2020towards} &\multirow{2}{*}{Mix} &2M& 13.2& 5.33& 0.199& 82.5 & 95.5& 98.5& 13.7& 8.01& 0.319& 83.2& 95.0& \underline{98.1} \\
			LeReS\textsuperscript{\dag}~ \cite{yin2021learning} &  & 354K& 14.1& 4.89& 0.190& 81.5 & 95.4& 98.3& 17.0& \underline{7.04}& 0.331& 76.1& 93.9 & 97.6 \\
			
			\midrule
			
			ZeroDepth~ \cite{guizilini2023towards} & \multirow{3}{*}{Mix} & 17M & 11.1 & 3.65 & 0.154 & 87.9 & 96.8 & 98.9 &\cellcolor{lightgray}8.62 &\cellcolor{lightgray}5.83 &\cellcolor{lightgray}0.156 &\cellcolor{lightgray}91.9 &\cellcolor{lightgray}97.4 &\cellcolor{lightgray}98.8 \\
			
			UniDepth~ \cite{piccinelli2024unidepth} &  & 3M & 13.6 & \textbf{3.13} & 0.149 & 88.0 & \textbf{99.3}& \textbf{99.9}& 13.0& \textbf{6.62}& \textbf{0.191}& \underline{84.4}& \textbf{97.1}& \textbf{98.9} \\
			
			Metric3D~ \cite{yin2023metric3d} & & 8M & \underline{7.91} & 3.34 & \underline{0.118} & \textbf{93.0} & \underline{98.9}& \underline{99.7}&\cellcolor{lightgray}12.0&\cellcolor{lightgray}6.52&\cellcolor{lightgray}0.178&\cellcolor{lightgray}87.7&\cellcolor{lightgray}94.5&\cellcolor{lightgray}98.1 \\

			\midrule
			\multirow{5}{*}{\textbf{GVDepth}} & KITTI & 23K &\cellcolor{lightgray}5.67 &\cellcolor{lightgray}2.61 &\cellcolor{lightgray}0.090&\cellcolor{lightgray}95.7&\cellcolor{lightgray}99.4&\cellcolor{lightgray}99.9&\textbf{11.8}&8.35&0.251&82.5&93.4&96.5\\
			
			& DDAD & 13K &12.3& 4.38 & 0.187 & 86.0 & 96.3 & 98.9&\cellcolor{lightgray}9.67&\cellcolor{lightgray}5.62&\cellcolor{lightgray}0.162&\cellcolor{lightgray}90.3&\cellcolor{lightgray}97.2&\cellcolor{lightgray}98.9\\
			
			& Argoverse & 5K & 11.2 & 4.43 & 0.181 & 86.1 & 97.1 & 99.4& 13.5& 8.38& 0.240& 82.6& 93.7& 96.9 \\
			
			& Waymo & 156K & 10.9 & 3.23 & 0.141 & 89.8 & 98.4 & 99.6&13.9& 7.68& 0.259& \textbf{86.2}& \underline{95.2}& 97.2 \\
			
			& DrivingStereo & 168K & \textbf{6.96} & \underline{3.17} & \textbf{0.110} & \underline{92.7} & 98.6 & 99.7&\underline{12.0} & 7.72 & \underline{0.215}& 82.4& 93.6& 97.6 \\
			
			\midrule
			
			UniDepth\textsuperscript{\ddag}~ \cite{piccinelli2024unidepth} & \multirow{2}{*}{Mix}& 3M & 5.61 & 2.34 & 0.084 & 97.0 & 99.6 & 99.9& 15.1& 5.82& 0.180& 89.3& 97.7& 99.0 \\
			
			Metric3D\textsuperscript{\ddag}~ \cite{yin2023metric3d} &  & 8M & 5.41 & 2.25 & 0.085 & 97.2 & 99.5 & 99.9&\cellcolor{lightgray}10.6&\cellcolor{lightgray}5.72&\cellcolor{lightgray}0.164&\cellcolor{lightgray}92.4&\cellcolor{lightgray}97.6&\cellcolor{lightgray}98.8 \\
			
			\bottomrule[0.4mm]	
	\end{tabular}}
	\label{tab:main}
	\vspace{-5pt}
\end{table*}
\noindent{}\textbf{Comparison with State-of-the-Art.}
We perform extensive evaluation, comparing GVDepth with SotA domain-dependent and domain-agnostic methods. Results for KITTI and DDAD datasets are shown in \cref{tab:main}.
Both supervised \cite{piccinelli2023idisc, yuan2022newcrfs} and self-supervised \cite{godard2019digging, zhou_diffnet} domain-dependent methods perform well on datasets matching their training domain but overfit to environmental and geometric biases, leading to reduced accuracy in KITTI $\rightarrow$ DDAD transfer.
Relative depth estimation models exhibit greater adaptability across domains; however, they require ground-truth scale and shift alignment, limiting their practical applicability.
Recent zero-shot metric methods, such as ZeroDepth, UniDepth, and Metric3D, demonstrate strong generalization capabilities but rely on extensive data acquisition and large-scale training, which is not always feasible.
Furthermore, UniDepth and Metric3D suffer accuracy degradation when inferring on resolutions lower than their high-resolution training data.
While resizing and padding can address this, these adjustments increase computational demands, limiting their suitability for resource-constrained systems.
GVDepth delivers comparable results while being adaptable to arbitrary camera setups, aspect ratios, and image resolutions.
Depending on the train-test dataset combination, our method ranks first or second across most metrics, despite training on only a single dataset and using a fraction of the data required by other zero-shot MDE methods.
This is particularly impressive, as each dataset features unique environments and camera setups, demonstrating that our model relies on domain-agnostic perspective geometry cues rather than domain-specific semantic cues.

\noindent{}\textbf{Comparison with Ground Plane Methods.}
\begin{table}[t]
	\centering
	\caption{\textbf{Comparison with GEDepth \cite{yang2023gedepth}.} Vertical -- depth regression via Vertical Canonical Transform - $\mathrm{VCT}_{\bm{\mathcal{C}}}(\cdot)$. Fusion -- depth regression with model as designed in \cref{fig:model}. All models are trained with equivalent setup and model complexity on KITTI.}
	\vspace{-5pt}
	\resizebox{1.0\linewidth}{!}{
		\begin{tabular}{cc|cc|cc|cc}
			\toprule[0.4mm]
			& \textbf{Testing} & \multicolumn{2}{c|}{\textbf{GEDepth}\cite{yang2023gedepth}}  & \multicolumn{2}{c|}{\textbf{Vertical}} & \multicolumn{2}{c}{\textbf{Fusion}} \\
			
			& \textbf{dataset} & $\mathrm{A.Rel}\downarrow$ & $\mathrm{\delta}_{1}\uparrow$ & $\mathrm{A.Rel}\downarrow$ & $\mathrm{\delta}_{1}\uparrow$ & $\mathrm{A.Rel}\downarrow$ & $\mathrm{\delta}_{1}\uparrow$ \\
			
			\midrule

			\multirow{5}{*}{\begin{turn}{90}\textbf{KITTI}\end{turn}}& KITTI & \cellcolor{lightgray}\underline{5.68} & \cellcolor{lightgray}95.4 & \cellcolor{lightgray}5.70 &\cellcolor{lightgray}\underline{95.5} &\cellcolor{lightgray}\textbf{5.67} & \cellcolor{lightgray}\textbf{95.7}\\ 
			& DStereo &18.25 &66.7 & \underline{10.43} & \underline{87.3} & \textbf{10.24} & \textbf{87.4} \\
			& Waymo & 22.11 & 69.1 & \textbf{13.42} & \textbf{78.6} & \underline{14.34} & \underline{78.4}\\
			& Argo & 19.27 & 52.8 & \underline{14.72} & \underline{64.7} & \textbf{10.45} &\textbf{84.8} \\
			& DDAD & 19.46 & 61.2 & \underline{14.26} & \underline{73.6} & \textbf{11.81} & \textbf{82.5} \\
					
			\bottomrule[0.4mm]
	\end{tabular}}
\label{tab:gedepth_small}
\vspace{-15pt}
\end{table}
Our proposed $\mathrm{VCT}_{\bm{\mathcal{C}}}(\cdot)$ implicitly encodes depth of the ground plane, which has been utilized in \cite{wang2023planedepth, yang2023gedepth, cecille2024groco} to improve MDE accuracy in autonomous driving scenes. 
However, as demonstrated in \cref{tab:gedepth_small}, $\mathrm{VCT}_{\bm{\mathcal{C}}}(\cdot)$ significantly outperforms GEDepth \cite{yang2023gedepth}, with the performance gap further widened after fusion with $\mathrm{FCT}_{\bm{\mathcal{C}}}(\cdot)$.
As we will show, the primary cause of such discrepency is inability of GEDepth to leverage vertical image positions and disentangle perspective geometry for pixels located outside the road surface.
In the Supp. materials we provide a comprehensive analysis and critique of existing models that rely on ground plane depth.

\noindent{}\textbf{Qualitative Results.}
\begin{figure}[t]
	\newcommand{\turnwidthnew}{0.47\columnwidth}
	\renewcommand{\arraystretch}{1}
	\centering
	\small
	\begin{tabular}{@{\hskip 0mm}r@{\hskip 1mm}c@{\hskip 1mm}c@{\hskip 0mm}}
		
		\raisebox{0.6\normalbaselineskip}[0pt][0pt]{\rotatebox{90}{Image}}& 
		\includegraphics[width=\turnwidthnew]{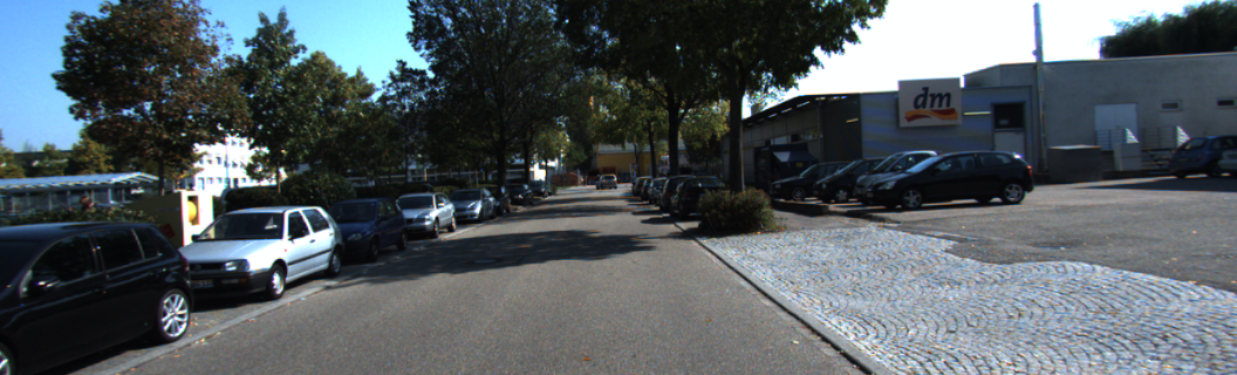} & \includegraphics[width=\turnwidthnew]{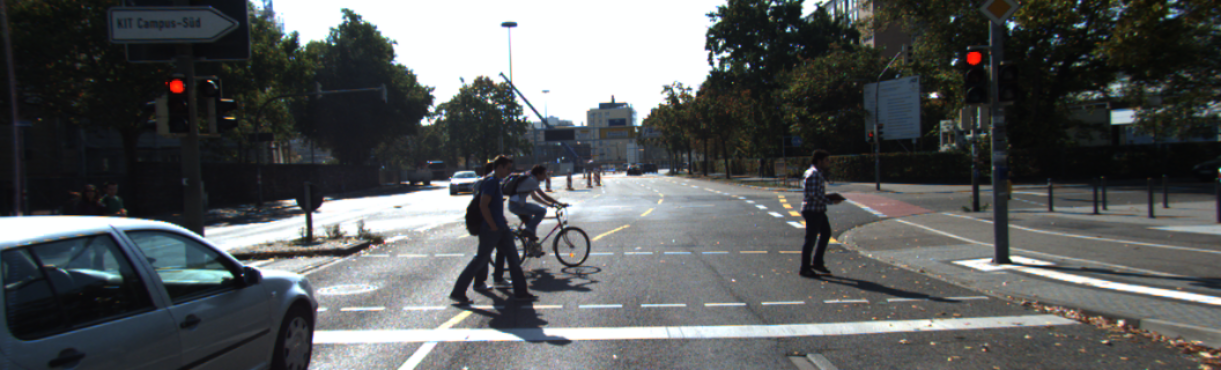}  \\
		
		\raisebox{1.0\normalbaselineskip}[0pt][0pt]{\rotatebox{90}{$\mathbf{D}$}} & 
		\includegraphics[width=\turnwidthnew]{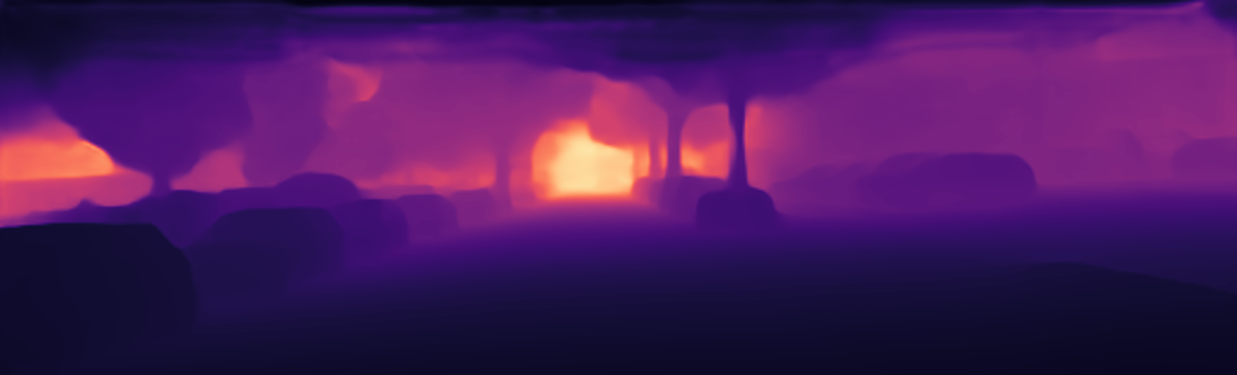} & \includegraphics[width=\turnwidthnew]{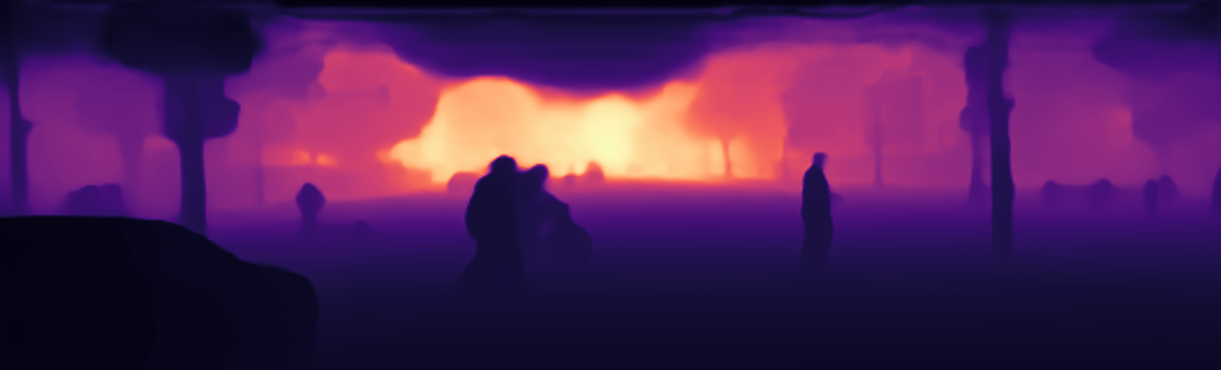}  \\
		\raisebox{0.5\normalbaselineskip}[0pt][0pt]{\rotatebox{90}{$\mathrm{A.Rel}$}} & 
		\includegraphics[width=\turnwidthnew]{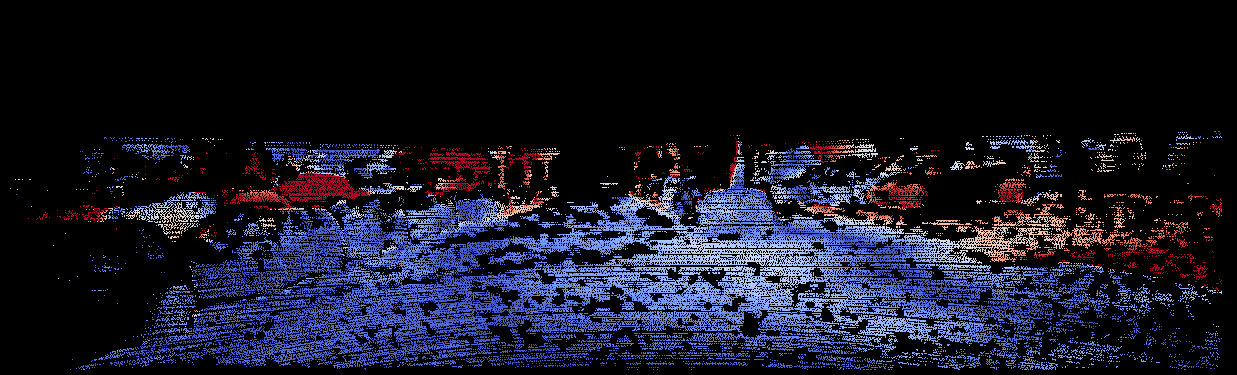} & \includegraphics[width=\turnwidthnew]{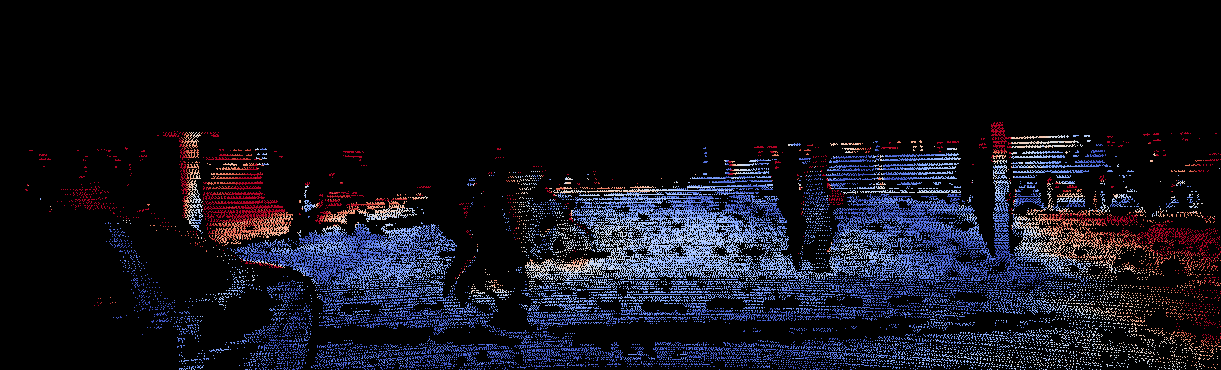} \\
		
		\raisebox{1.0\normalbaselineskip}[0pt][0pt]{\rotatebox{90}{$\mathbf{\Sigma_F}$}} & 
		\includegraphics[width=\turnwidthnew]{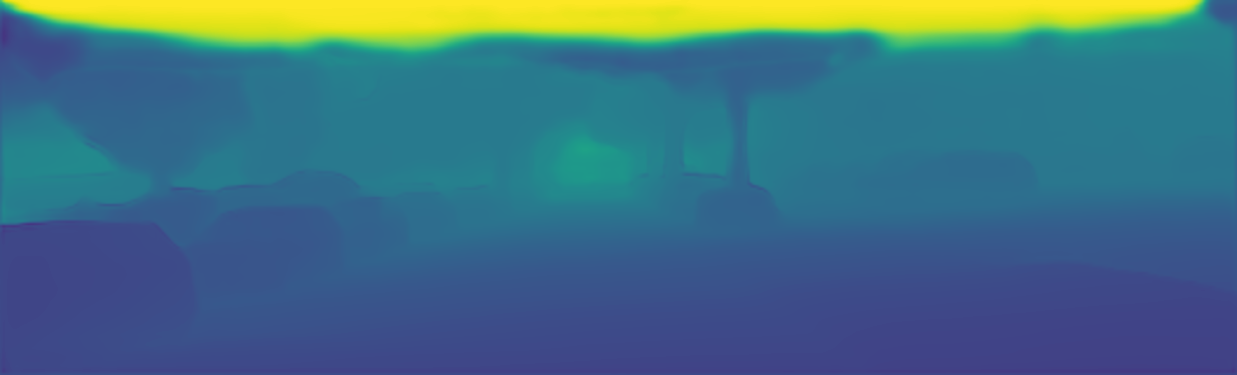} & \includegraphics[width=\turnwidthnew]{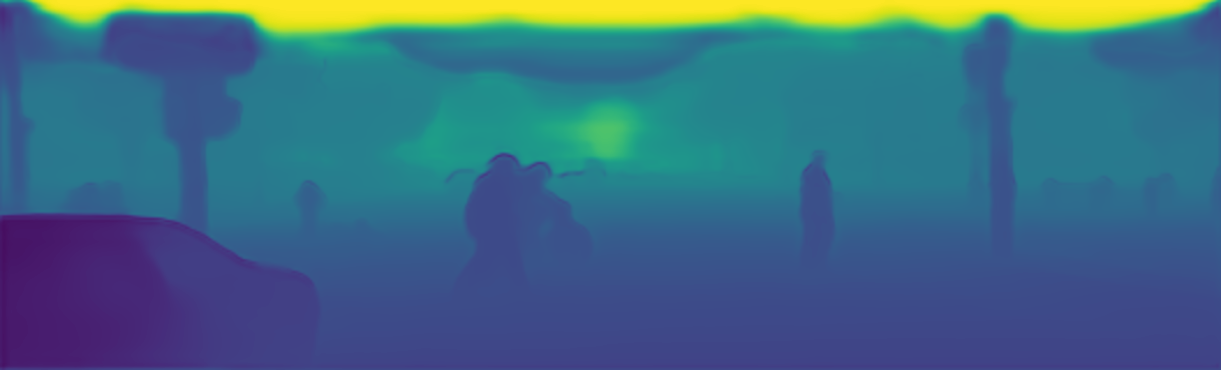}\\
		\raisebox{1.0\normalbaselineskip}[0pt][0pt]{\rotatebox{90}{$\mathbf{\Sigma_Y}$}} & 
		\includegraphics[width=\turnwidthnew]{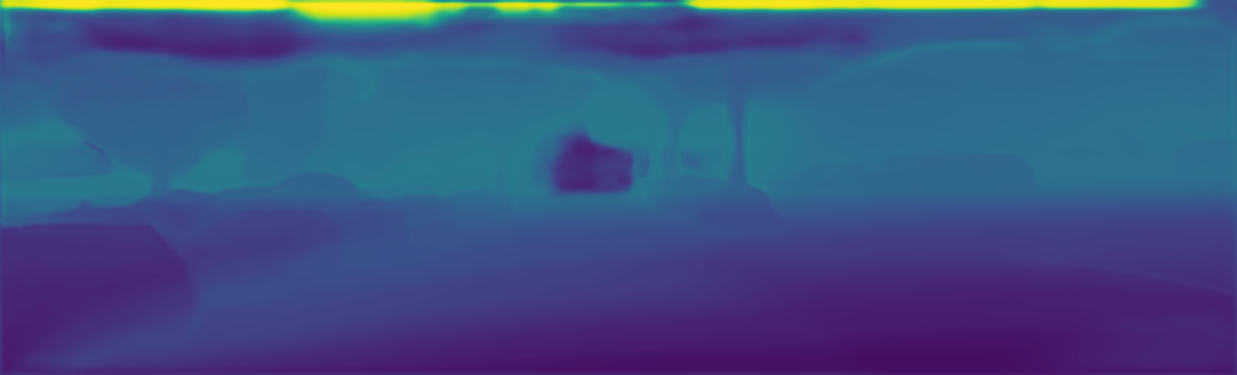} & \includegraphics[width=\turnwidthnew]{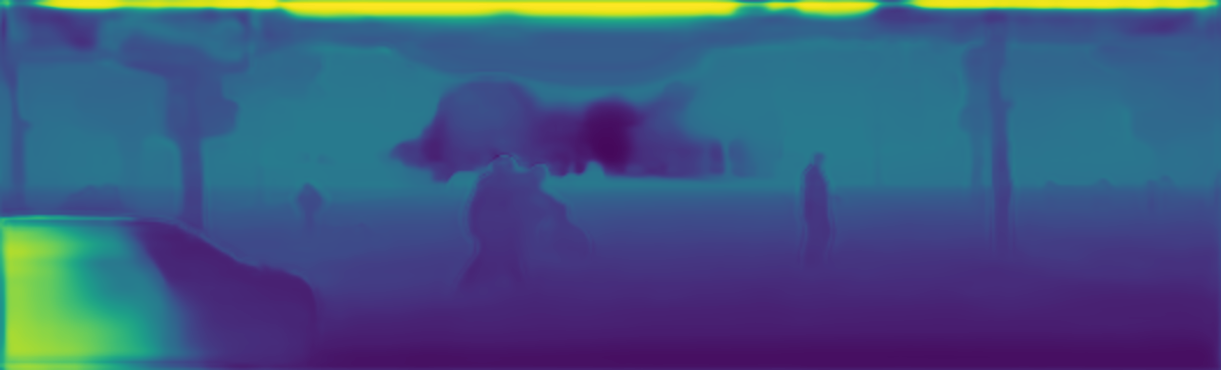}\\
		
		 & 
		\includegraphics[width=\turnwidthnew]{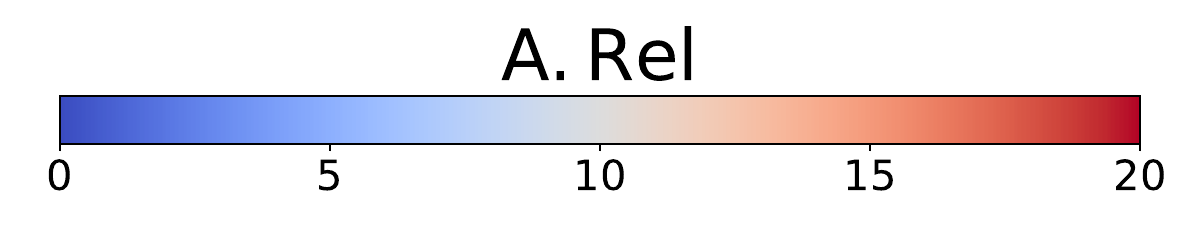} & \includegraphics[width=\turnwidthnew]{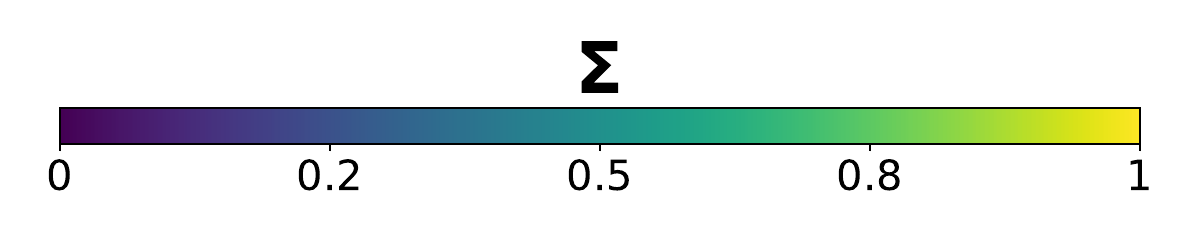}\\
	\end{tabular}
	\vspace{-8pt}
	\caption{\textbf{Qualitative results.} Predicted depths, errors and uncertanties for DrivingStereo $\rightarrow$ KITTI evaluation.}
	\label{fig:uncertainty}
	\vspace{-10pt}
\end{figure}
In \cref{fig:uncertainty} we illustrate qualitative results of our DrivingStereo $\rightarrow$ KITTI evaluation.
Notably, it can be seen that vertical uncertainty $\mathbf{\Sigma_Y}$ is lower in closer regions, and decreasing with higher distances.
An interesting detail appears in the left image, where the vertical uncertainty adapts to the scene’s leftward slope, indicating that the model is aware of ground perturbations.
Additionally, in the right image, the vertical uncertainty is significantly higher for a car lacking a visible ground-contact point, creating ambiguity in its vertical image position cue.

\subsection{Ablation studies}
\begin{table*}[t]
	\centering
	\caption{\textbf{Ablation of canonical representations.} Baseline -- standard depth regression with geometric augmentations. Focal -- depth regression via Focal Canonical Transform - $\mathrm{FCT}_{\bm{\mathcal{C}}}(\cdot)$. This is equivalent to Metric3D\cite{yin2023metric3d}, but trained on our setup. Vertical -- depth regression via Vertical Canonical Transform - $\mathrm{VCT}_{\bm{\mathcal{C}}}(\cdot)$. Fusion -- depth regression with model as designed in \cref{fig:model}. Training datasets on rows, testing datasets on columns. Best results are \textbf{bolded}, second best are \underline{underlined}. In-domain evaluation results are \colorbox{lightgray}{shaded}. \textbf{Note:} Focal and vertical results presented here are from separate models leveraging  $\text{FCT}_{\bm{\mathcal{C}}}(\cdot)$ and  $\text{VCT}_{\bm{\mathcal{C}}}(\cdot)$, not from submodels in \cref{fig:model}.}
	\vspace{-5pt}
	\resizebox{\linewidth}{!}{
		\begin{tabular}{cl|c@{\hspace{6pt}}cc|c@{\hspace{6pt}}cc|c@{\hspace{6pt}}cc|c@{\hspace{6pt}}cc|c@{\hspace{6pt}}cc}
			
    		\toprule[0.4mm]
			& \multirow{2}{*}{\textbf{Representation}} & \multicolumn{3}{c|}{\textbf{KITTI}} & \multicolumn{3}{c|}{\textbf{DrivingStereo}} & \multicolumn{3}{c|}{\textbf{Waymo}} & \multicolumn{3}{c|}{\textbf{Argoverse}} & \multicolumn{3}{c}{\textbf{DDAD}} \\
			& & $\mathrm{A.Rel}\downarrow$ & \phantom{x}$\mathrm{RMS}\downarrow$ & $\mathrm{\delta}_{1}\uparrow$ & $\mathrm{A.Rel}\downarrow$ & \phantom{x}$\mathrm{RMS}\downarrow$ & $\mathrm{\delta}_{1}\uparrow$ & $\mathrm{A.Rel}\downarrow$ & \phantom{x}$\mathrm{RMS}\downarrow$ & $\mathrm{\delta}_{1}\uparrow$ & $\mathrm{A.Rel}\downarrow$ & \phantom{x}$\mathrm{RMS}\downarrow$ & $\mathrm{\delta}_{1}\uparrow$ & $\mathrm{A.Rel}\downarrow$ & \phantom{x}$\mathrm{RMS}\downarrow$ & $\mathrm{\delta}_{1}\uparrow$ \\
			\midrule
			\multirow{4}{*}{\rotatebox{90}{\textbf{KITTI}}} 
			& Baseline & \cellcolor{lightgray}6.01 & \cellcolor{lightgray}2.65 & \cellcolor{lightgray}95.2 &25.60 &8.31 &23.0 &31.61 &10.86 &11.0 &46.28 &15.33 &6.0 &16.12 &8.72 &76.3 \\
			& Focal & \cellcolor{lightgray}5.78 & \cellcolor{lightgray}2.78 & \cellcolor{lightgray}94.8 & 12.01 & 5.77 & 86.1 & 19.12 & 8.30 & 72.5 & \underline{10.51} & \textbf{4.86} & \textbf{86.6} & \underline{14.15} & \underline{8.41} & \underline{80.0} \\
			& Vertical & \cellcolor{lightgray}\underline{5.70} & \cellcolor{lightgray}\textbf{2.58} & \cellcolor{lightgray}\underline{95.5} & \underline{10.43} & \textbf{5.42} & \underline{87.3} & \textbf{13.42} & \textbf{7.46} & \textbf{78.6} & 14.72 & 7.11 & 64.7 & 14.26 & 9.39 & 73.6 \\
			& Fusion & \cellcolor{lightgray}\textbf{5.67} & \cellcolor{lightgray}\underline{2.61} & \cellcolor{lightgray}\textbf{95.7} & \textbf{10.24} & \underline{5.66} & \textbf{87.4} & \underline{14.34} & \underline{7.56} & \underline{78.4} & \textbf{10.45} & \underline{5.06} & \underline{84.8} & \textbf{11.81} & \textbf{8.35} & \textbf{82.5} \\
			\midrule
			\multirow{4}{*}{\rotatebox{90}{\textbf{DStereo}}} 
			& Baseline & 35.62 & 5.56& 20.1& \cellcolor{lightgray}4.82 & \cellcolor{lightgray}2.76 & \cellcolor{lightgray}97.9 &20.21 &9.02 & 65.1&20.62 &8.51 &50.7 &25.32 &8.10 &47.7 \\
			& Focal & 8.47 & 3.74 & 89.8 & \cellcolor{lightgray}\underline{3.05} & \cellcolor{lightgray}1.76 & \cellcolor{lightgray}99.5 & 16.61 & 8.04 & 74.2 & \underline{9.94} & \underline{4.95} & \underline{86.8} & 14.92 & \underline{7.86} & \underline{82.0} \\
			& Vertical & \underline{7.42} & \underline{3.33} & \underline{92.6} & \cellcolor{lightgray}3.07 & \cellcolor{lightgray}\textbf{1.75} & \cellcolor{lightgray}\underline{99.5} & \textbf{11.82} & \textbf{6.36} & \textbf{85.5} & 11.41 & 6.56 & 84.7 & \underline{14.71} & 8.42 & 80.0 \\
			& Fusion & \textbf{6.96} & \textbf{3.17} & \textbf{92.7} & \cellcolor{lightgray}\textbf{3.01} & \cellcolor{lightgray}\underline{1.76} & \cellcolor{lightgray}\textbf{99.5} & \underline{12.15} & \underline{6.50} & \underline{83.1} & \textbf{9.91} & \textbf{4.92} & \textbf{86.9} & \textbf{12.02} & \textbf{7.72} & \textbf{82.4} \\
			\midrule
			\multirow{4}{*}{\rotatebox{90}{\textbf{Waymo}}} 
			& Baseline & 18.32&5.65 &84.0 &22.50 & 5.98&64.9 & \cellcolor{lightgray}4.43 &2.86 \cellcolor{lightgray} & \cellcolor{lightgray}96.8 &28.74 &9.51 &10.9 &24.33 &8.92 & 55.1\\
			& Focal & 18.21 & 5.87 & 80.2 & 13.81 & \underline{5.27} & 85.5 & \cellcolor{lightgray}3.62 & \cellcolor{lightgray}\textbf{2.46} & \cellcolor{lightgray}98.6 & 18.41 & 5.87 & 85.3 & 20.41 & 8.75 & 66.0 \\
			& Vertical & \textbf{8.30} & \textbf{3.17} & \textbf{92.3} & \textbf{11.40} & \textbf{5.25} & \underline{86.8} & \cellcolor{lightgray}\underline{3.61} & \cellcolor{lightgray}\underline{2.48} & \cellcolor{lightgray}\textbf{98.9} & \underline{12.21} & \underline{5.12} & \underline{88.6} & \textbf{11.51} & \textbf{7.34} & \underline{84.4} \\
			& Fusion & \underline{10.92} & \underline{3.23} & \underline{89.8} & \underline{12.78} & 5.68 & \textbf{87.1} & \cellcolor{lightgray}\textbf{3.51} & \cellcolor{lightgray}2.57 & \cellcolor{lightgray}\underline{98.8} & \textbf{9.62} & \textbf{4.74} & \textbf{94.1} & \underline{13.99} & \underline{7.68} & \textbf{86.2} \\
			
			\midrule
			
			\multirow{4}{*}{\begin{turn}{90}\textbf{Argo}\end{turn}} &Baseline &72.31 & 9.91 &7.6 & 46.32 & 9.99 & 20.2 &34.41 &8.71 &34.8&\cellcolor{lightgray}4.72&\cellcolor{lightgray}2.45&\cellcolor{lightgray}97.6&41.92&9.35&25.4 \\
			&Focal & \underline{13.16} & 6.41 & 80.2 & \underline{16.61} & 10.21 & 74.6 & 21.13 & 9.40 &60.6&\cellcolor{lightgray}3.16&\cellcolor{lightgray}\underline{1.81}&\cellcolor{lightgray}98.6&18.21&10.51& 62.6\\
			&Vertical & 14.60& \underline{5.81} & \underline{84.2} & 18.22 & \underline{8.10} & \underline{75.6} &\underline{14.98} & \textbf{7.38} &\underline{72.6}&\cellcolor{lightgray}\textbf{2.95}&\cellcolor{lightgray}\textbf{1.77}&\cellcolor{lightgray}\underline{98.6}&\underline{13.86}&\underline{8.88}& \underline{81.5} \\
			&Fusion & \textbf{11.21}& \textbf{4.43} & \textbf{86.1} & \textbf{12.81} & \textbf{7.81} & \textbf{82.1} & \textbf{12.92}& \underline{7.38}&\textbf{78.4}&\cellcolor{lightgray}\underline{3.14}&\cellcolor{lightgray}1.82&\cellcolor{lightgray}\textbf{98.7}&\textbf{13.48}&\textbf{8.38}& \textbf{82.6}\\

			\midrule
			
			\multirow{4}{*}{\begin{turn}{90}\textbf{DDAD}\end{turn}} &Baseline &59.42&10.17&5.6&18.61&6.72&77.0&26.61&8.40&30.9&31.81&9.98&15.1&10.98\cellcolor{lightgray}&\cellcolor{lightgray}5.86&\cellcolor{lightgray}88.2\\
			&Focal & \underline{12.45} & \textbf{4.07} & \underline{86.0} & \underline{14.31} & \textbf{5.64} & \underline{83.1} & 21.91& 7.60&53.9&15.01&6.16&83.1&\cellcolor{lightgray}10.91&\cellcolor{lightgray}5.63&\cellcolor{lightgray}88.7\\
			&Vertical &22.21 & 6.17 & 72.5 & 17.00 & 6.45 & 79.1 & \textbf{15.51}&\textbf{6.27}&\textbf{80.6}&\underline{14.10}& \underline{5.96}&\underline{83.1}&\cellcolor{lightgray}\underline{10.20}&\cellcolor{lightgray}\underline{5.73}&\cellcolor{lightgray}\underline{89.1}\\
			&Fusion &\textbf{12.31} & \underline{4.38} &\textbf{86.0} & \textbf{14.21} & \underline{6.31} & \textbf{83.8} & \underline{16.61}& \underline{7.36}&\underline{77.6}&\textbf{12.66}&\textbf{5.76}&\textbf{84.9}&\cellcolor{lightgray}\textbf{9.67}&\cellcolor{lightgray}\textbf{5.62}&\cellcolor{lightgray}\textbf{90.3}\\
			\bottomrule[0.4mm]
	\end{tabular}}

	\label{tab:ablation}
	\vspace{-4pt}
\end{table*}
\noindent{}\textbf{Evaluation of Canonical Representations.}
In \cref{tab:ablation} we report extensive evaluation results, spanning different representations and train-test dataset combinations. Moreover, in Supplementary we report qualitative results accompanying the table.
By carefully examining the results, we can draw the following conclusions: i) \textbf{The baseline model struggles to transfer effectively to out-of-distribution data.} 
Without the invariance introduced by canonical representations, baseline models fail to leverage diverse perspective geometries in the training data due to the ambiguities induced by depth-camera parameters entanglement; (ii) \textbf{Vertical image position cues generalize better than object size cues.} 
Models with our novel $\text{VCT}_{\bm{\mathcal{C}}}(\cdot)$ representation generally outperform $\text{FCT}_{\bm{\mathcal{C}}}(\cdot)$ models, except in rare cases, potentially stemming from calibration errors.
This is particularly evident with models trained on the Waymo dataset, which encounter difficulties when relying solely on object size cues.
The probable cause is that Waymo's cameras are positioned significantly higher, causing a variation of apparent object sizes due to the change of perspective.
In contrast, vertical image position cues are more robust, implicitly encoding camera height within the representation; (iii) \textbf{Probabilistic fusion improves generalization by integrating both cues.} Our fusion strategy achieves the most accurate predictions by adaptively weighting the more reliable cue.
For example,  when object size cues are more accurate (\eg DDAD $\rightarrow$ KITTI), the model adjusts accordingly, and the same applies for vertical position cues (\eg Waymo $\rightarrow$ Argoverse).

\begin{figure}
	\includegraphics[width=\linewidth, trim=40pt 0pt 40pt 40pt, clip]{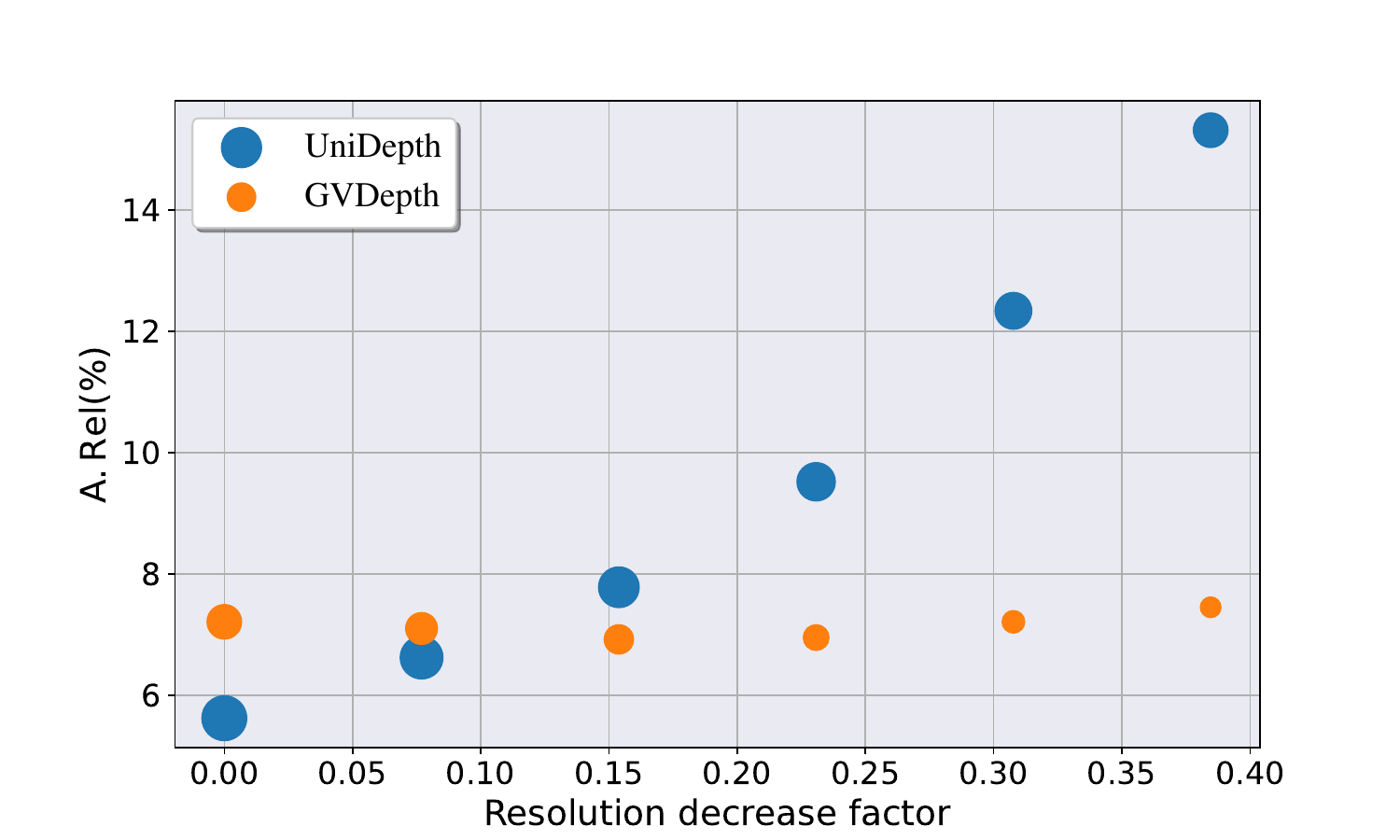}
	\vspace{-6pt}
	\centering
	\caption{\textbf{Resolution adaptability.} $\mathrm{A.Rel} (\%)$ of UniDepth \cite{piccinelli2024unidepth} and GVDepth as height resolution $H$ decreases from base resolution 462. Computational complexity is depicted with different circle radius. GVDepth results are from DrivingStereo $\rightarrow$ KITTI evaluation.}
	\label{fig:resolution}
	\vspace{-14pt}
\end{figure}
\noindent{}\textbf{Resolution Adaptability.}
SoTA zero-shot methods such as UniDepth \cite{piccinelli2024unidepth} and Metric3D \cite{yin2023metric3d} require resizing and padding to training resolution to achieve optimal accuracy.
However, when applied to datasets like KITTI, this approach results in the model processing nearly twice as many pixels as necessary due to padding.
This significantly increases computational complexity without adding any meaningful detail to the image.

In contrast, GVDepth is designed to be fully resolution-agnostic, adapting seamlessly to the native resolution of the input image.
As shown in \cref{fig:resolution}, the error rate of GVDepth remains largely unaffected even as the input resolution decreases.
This feature is especially advantageous for real-time systems, where dynamically adjusting image resolution is one of the simplest and most effective ways to control computational complexity.

\begin{table}[!t]
	\centering

	\caption{\textbf{Surround-view generalization.} Vertical -- model with $\mathrm{VCT}_{\bm{\mathcal{C}}}(\cdot)$. Focal -- model with $\mathrm{FCT}_{\bm{\mathcal{C}}}(\cdot)$. All models evaluated on DDAD. $\mathrm{VCT}_{\bm{\mathcal{C}}}(\cdot)$ and $\mathrm{FCT}_{\bm{\mathcal{C}}}(\cdot)$ models trained solely on Waymo. (\dag): Waymo surround images added to training data. UniDepth and Metric3D evaluated with original checkpoints and resolutions.}
	\vspace*{-8pt}
	\resizebox{\linewidth}{!}{
	\begin{tabular}{l|cc|cc|cc}
		\toprule[0.4mm]
		\multirow{2}{*}{Method} & \multicolumn{2}{c|}{Front-left}& \multicolumn{2}{c|}{Front-left} & \multicolumn{2}{c}{Back}
		\\
		&
		$\mathrm{A.Rel}\downarrow$ & $\mathrm{\delta}_{1}\uparrow$ & $\mathrm{A.Rel}\downarrow$ & $\mathrm{\delta}_{1}\uparrow$ & $\mathrm{A.Rel}\downarrow$ & $\mathrm{\delta}_{1}\uparrow$ \\
		
		\midrule
		Baseline  & 67.42 & 18.2 & 78.48 & 12.3 & 33.58 & 27.1\\
		Focal & 23.37 & 62.2 & 22.70 & 67.1 & \underline{20.93} & \underline{67.7}\\
		Vertical & \underline{19.21} & \underline{71.8} & \underline{17.22} & \textbf{73.5} & 22.12 & 66.2\\
		Focal\dag &  20.69 & 68.2& 20.01 & 66.7 &\textbf{19.60} & \textbf{68.6} \\
		Vertical\dag & \textbf{16.44} & \textbf{76.5} & \textbf{16.82} & \underline{71.2}&  21.35 & 65.8\\
		\midrule
		UniDepth\dag ~ \cite{piccinelli2024unidepth} & 72.28 & 8.9 & 71.40 & 8.2 &  55.51 & 11.1 \\
		
		Metric3D\dag ~ \cite{yin2023metric3d}  &\cellcolor{lightgray}14.57 & \cellcolor{lightgray}80.7 &\cellcolor{lightgray}13.86 & \cellcolor{lightgray}80.1 &\cellcolor{lightgray}10.55 & \cellcolor{lightgray}88.7 \\
		\bottomrule[0.4mm]
	\end{tabular}}
	\label{tbl:surroundview}
	\vspace{-0pt}
\end{table}

\begin{figure}[t]
	\includegraphics[width=0.95\linewidth, trim=0 15pt 0 10pt, clip]{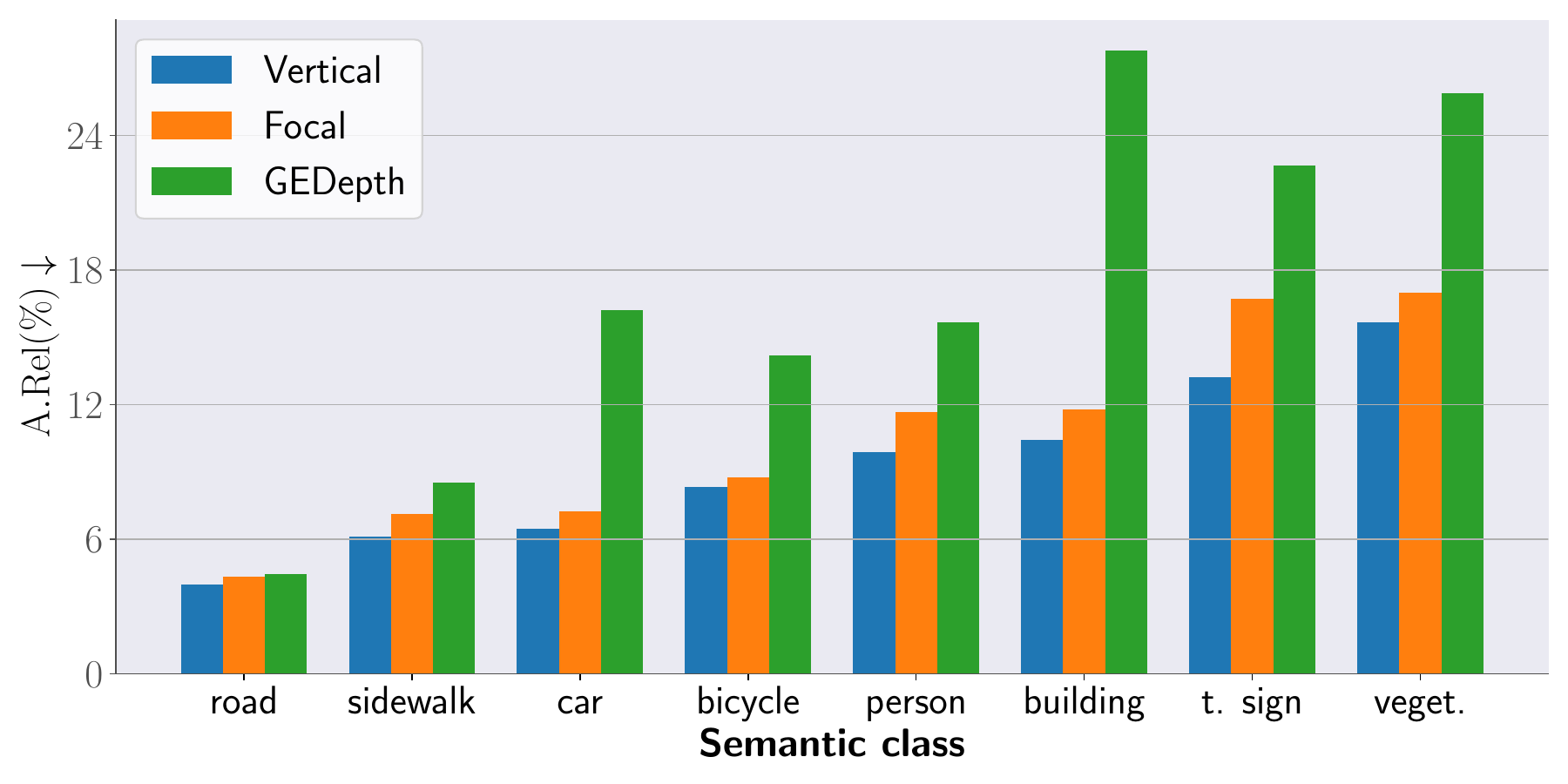}
	\caption{\textbf{Error per object class.} Vertical -- model with $\mathrm{VCT}_{\bm{\mathcal{C}}}(\cdot)$. Focal -- model with $\mathrm{FCT}_{\bm{\mathcal{C}}}(\cdot)$. All models are evaluated on DrivingStereo $\rightarrow$ KITTI zero-shot transfer.}
	\label{fig:semantics}
	\vspace{-8pt}
\end{figure}
\noindent\textbf{Surround-view Generalization.} In \cref{tbl:surroundview} we provide results for Waymo $\to$ DDAD zero-shot transfer for surround-view cameras. Despite the increased ambiguity and perturbations in ground regions captured by side views, our $\operatorname{VCT(\cdot)}$ model outperforms $\operatorname{FCT(\cdot)}$ on side cameras, with performance increasing after Waymo side cameras are included in training data.
$\operatorname{FCT(\cdot)}$ performs better on back camera, possibly due to the occlusion of bottom part of the image by ego-vehicle.
For unknown reasons, UniDepth \cite{piccinelli2024unidepth} failed to infer correct camera parameters for DDAD surround cameras, resulting in subpar performance.

\noindent{}\textbf{Generalization in Specific Regions.}
The strong quantitative generalization results of vertical position cues do not fully capture their performance.
For instance, a model with  $\mathrm{VCT}_{\bm{\mathcal{C}}}(\cdot)$ might only perform better on road pixels due to its implicitly encoded planarity.
However, as shown in \cref{fig:semantics}, this is not the case: the $\mathrm{VCT}_{\bm{\mathcal{C}}}(\cdot)$ model exhibits accurate generalization across a wide range of object classes. The results clearly reveal major flaws in GEDepth's \cite{yang2023gedepth} performance, as it struggles to generalize beyond road-related pixels, particularly for safety-critical objects.

\begin{figure}[t]
	\centering
	\begin{subfigure}[t]{0.5\linewidth}
		\centering
		\includegraphics[width=\linewidth]{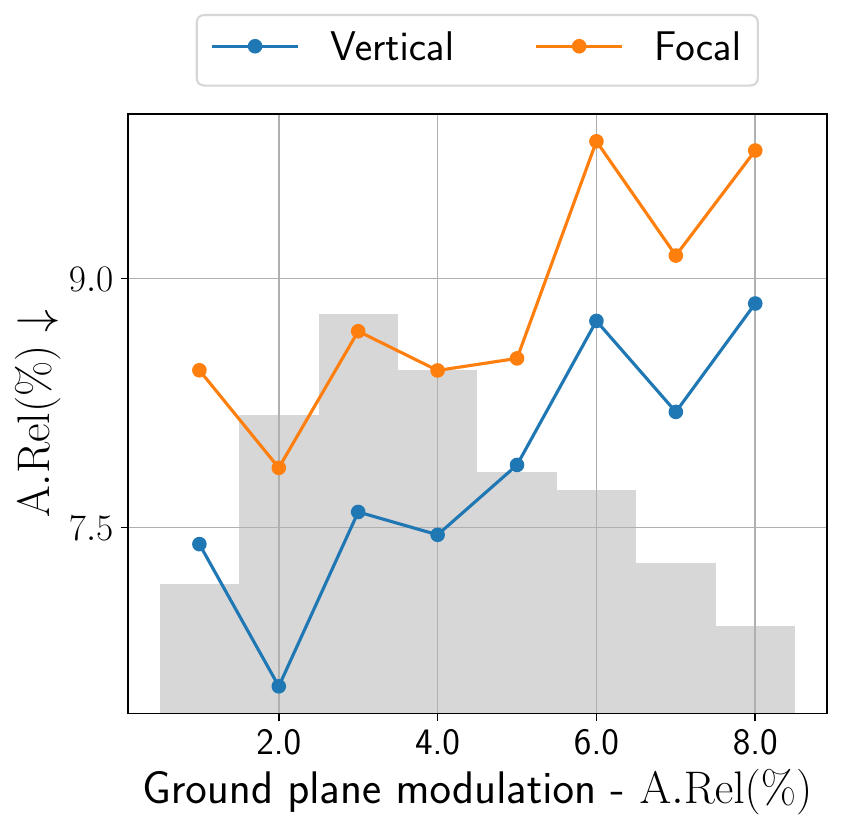} 
		\caption{}
		\label{fig:modulation}
	\end{subfigure}%
	\hfill
	\begin{subfigure}[t]{0.5\linewidth}
		\centering
		\includegraphics[width=\linewidth]{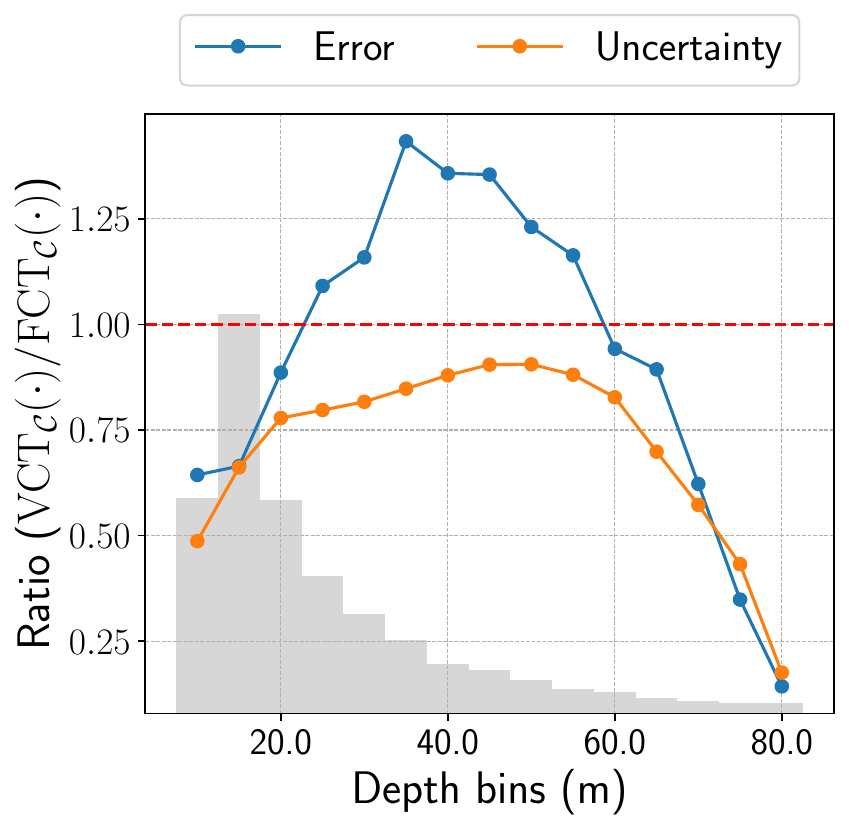} 
		\caption{}
		\label{fig:ratio}
	\end{subfigure}
	\vspace{-5pt}
	\caption{\textbf{Sensitivity analysis.} All models are evaluated on DrivingStereo $\rightarrow$ KITTI. Histograms represent data distribution. \textbf{a)} $\mathrm{A.Rel}$ of $\mathrm{VCT}_{\bm{\mathcal{C}}}(\cdot)$ and $\mathrm{FCT}_{\bm{\mathcal{C}}}(\cdot)$ models depending on ground plane modulation severity. Ground plane modulation is expressed as $\mathrm{A.Rel}$ of ground depth calculated from camera parameters via \cref{eqn:vertical} and ground-truth depth of the road. \textbf{b)} Ratio of depth errors $(\mathrm{A.Rel}(\mathbf{Z_Y}) / \mathrm{A.Rel}(\mathbf{Z_F}))$ and uncertainties $(\mathbf{\Sigma}_Y/\mathbf{\Sigma}_Z)$ depending on depth bin.}
	\label{fig:_}
	\vspace{-10pt}
\end{figure}

Moreover, we examine the effects of ground plane inconsistencies and perturbations on the depth estimation accuracy.
A reasonable assumption would be that the $\mathrm{VCT}_{\bm{\mathcal{C}}}(\cdot)$ model is more sensitive to these perturbations.
However, \cref{fig:modulation} illustrates that error for both models increases in similar pattern as ground perturbation increases.
This indicates that our $\mathrm{VCT}_{\bm{\mathcal{C}}}(\cdot)$ model can adapt and recognize perturbations which are present in the training data, atleast to the similar degree as competing methods.

Finally, in \cref{fig:ratio} we analyze the errors and uncertainties of both cues across discrete depth bins.
As shown, the error ratio skews towards $\mathrm{VCT}_{\bm{\mathcal{C}}}(\cdot)$ in regions closer to the camera, which is expected.
As depth increases, the object size cue becomes more accurate due to the reduced reliability of the vertical position cue at mid-range distances.
However, for farther distances, $\mathrm{VCT}_{\bm{\mathcal{C}}}(\cdot)$ once again achieves lower error.
We hypothesize that this is because $\mathrm{VCT}_{\bm{\mathcal{C}}}(\cdot)$ implicitly encodes the horizon level, aiding depth estimation for distant objects.
Additionally, the estimated uncertainties follow a similar pattern, demonstrating the effectiveness of our adaptive fusion module.

\section{Conclusion and Future Work}
\label{sec:conclusion}
We present GVDepth, a novel MDE model designed for accurate zero-shot transfer across diverse environments and camera setups, with a focus on autonomous vehicles and mobile robotics. Leveraging the fixed camera-to-ground geometry in such platforms, we introduce a \emph{Vertical Canonical Representation} that uses known ground depth to enforce perspective invariance, resulting in enhanced learning and generalization.
Our architecture further adaptively fuses depth estimates from object size and vertical image position cues, guided by uncertainty.

Extensive ablations across five autonomous driving datasets show that our method generalizes significantly better than object-size-based baselines, matching or surpassing state-of-the-art zero-shot models, while using only a single training dataset and far fewer computational and data resources.
Future work will explore scaling to multi-dataset training toward a foundation model for MDE in autonomous driving and mobile robotics.
\def\thesection{\Alph{section}}

\twocolumn[{%
	\centering
	\textbf{\Large  Supplementary Material }
	\vspace{20pt}
	\begin{center}
	\newcommand{\turnheightnew}{38pt}
	\newcommand{\colorbarheight}{182pt}
	\renewcommand{\arraystretch}{1}
	\centering
	\small
	\captionsetup{type=figure}
	\begin{tabular}{@{\hskip 0mm}r@{\hskip 1mm}c@{\hskip 1mm}c@{\hskip 2.5mm}c@{\hskip 1mm}c@{\hskip 2.5mm}c@{\hskip 1mm}c@{\hskip 0mm}c@{\hskip 0mm}}
		
		& \multicolumn{2}{c}{\hspace{-10pt}KITTI}  & \multicolumn{2}{c}{\hspace{-10pt}Waymo} & \multicolumn{2}{c}{DDAD} &\\
		
		& \multicolumn{2}{c}{\hspace{-10pt}\includegraphics[height=\turnheightnew, trim=120pt 0 120pt 0, clip]{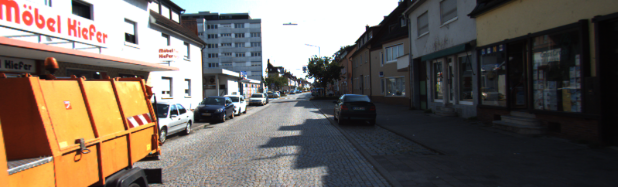}}  & \multicolumn{2}{c}{\hspace{-10pt}\includegraphics[height=\turnheightnew]{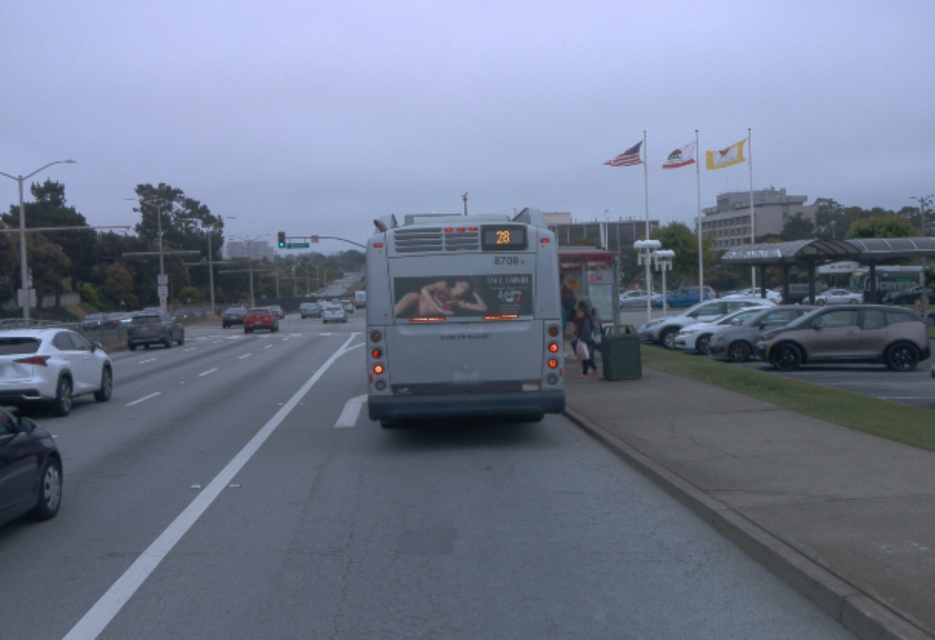}} & \multicolumn{2}{c}{\includegraphics[height=\turnheightnew]{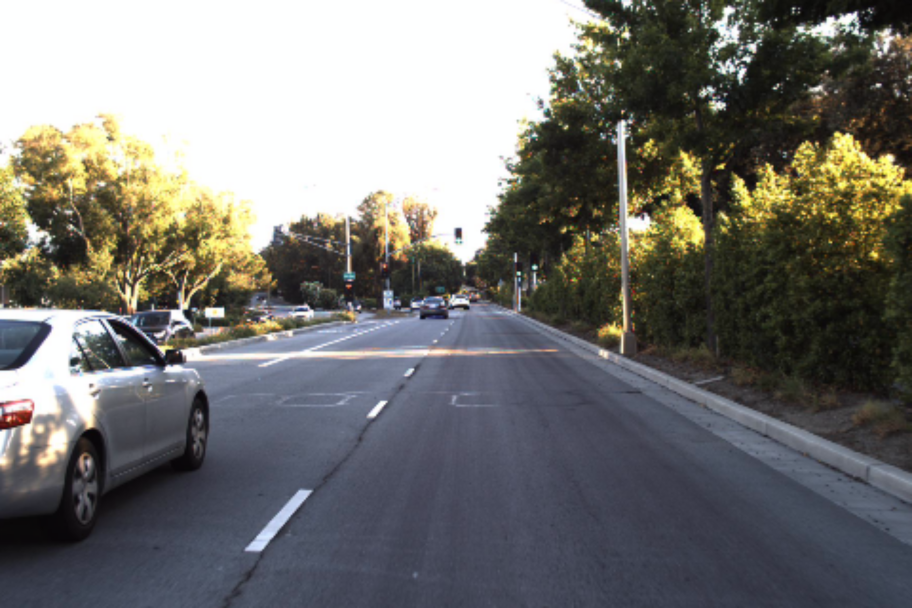}} & \multirow{4}{*}{\includegraphics[height=\colorbarheight]{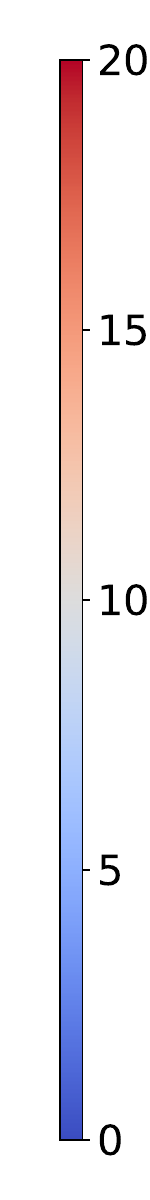}}\\
		
		\raisebox{0.3\normalbaselineskip}[0pt][0pt]{\rotatebox{90}{Baseline}}& 
		\includegraphics[height=\turnheightnew, trim=120pt 0 120pt 0, clip]{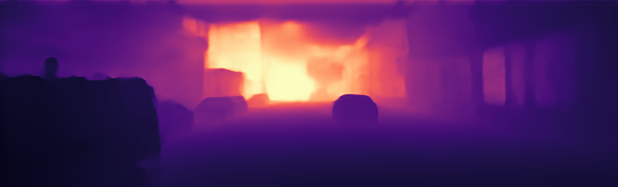} & \includegraphics[height=\turnheightnew, trim=120pt 0 120pt 0, clip]{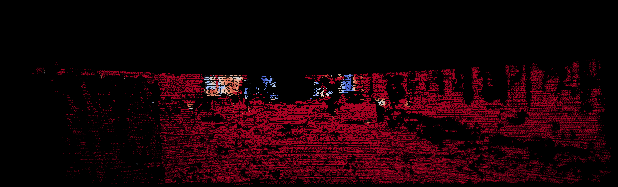} &
		\includegraphics[height=\turnheightnew]{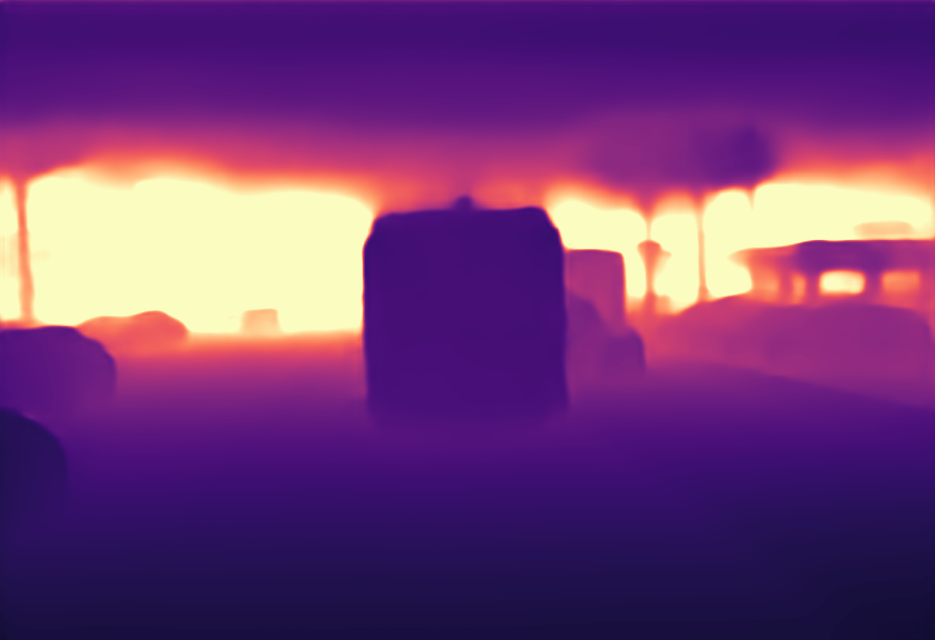} & \includegraphics[height=\turnheightnew]{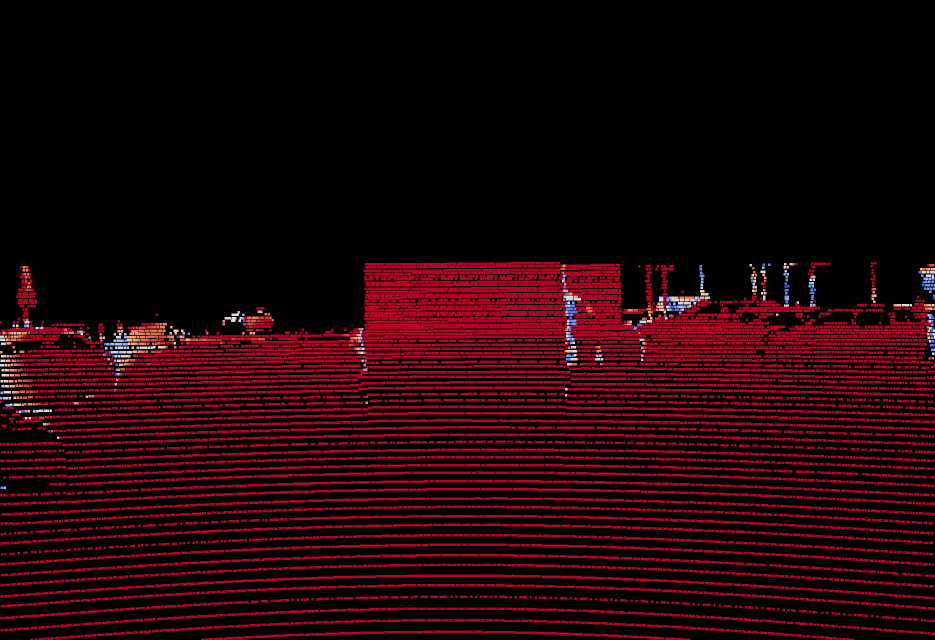} &
		\includegraphics[height=\turnheightnew]{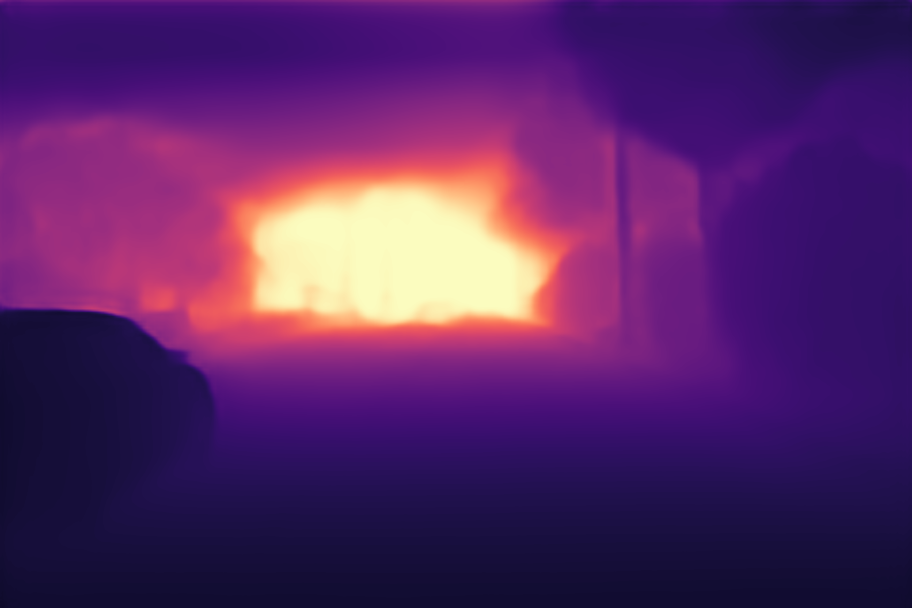} & \includegraphics[height=\turnheightnew]{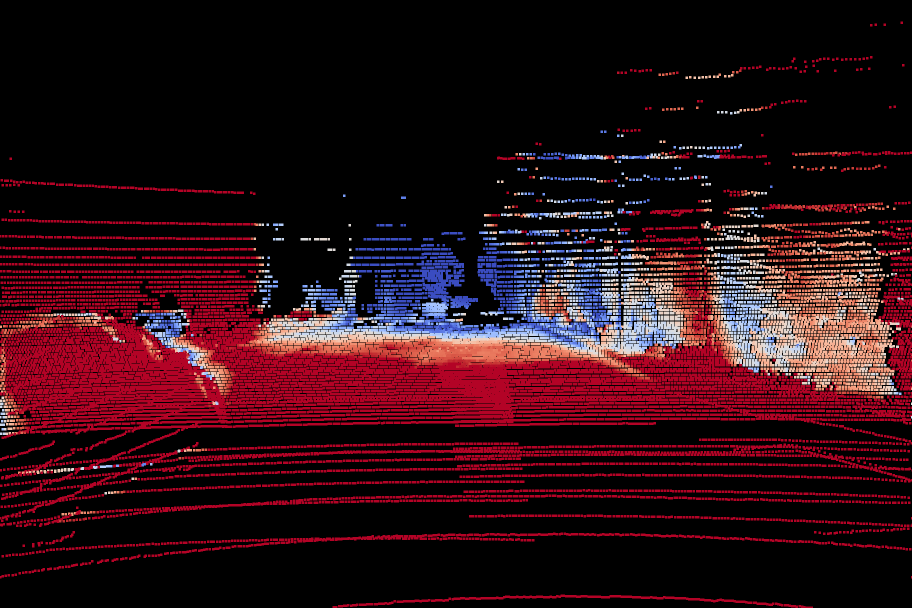}&\\
		
		\raisebox{0.8\normalbaselineskip}[0pt][0pt]{\rotatebox{90}{Focal}}& 
		\includegraphics[height=\turnheightnew, trim=120pt 0 120pt 0, clip]{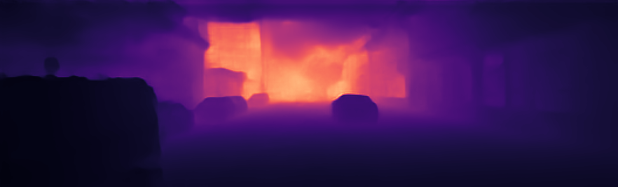} & \includegraphics[height=\turnheightnew, trim=120pt 0 120pt 0, clip]{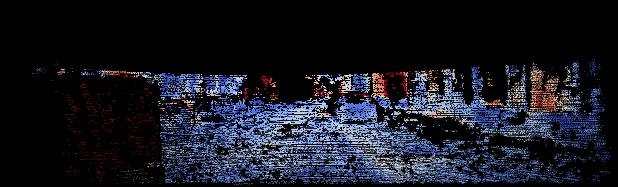} &
		\includegraphics[height=\turnheightnew]{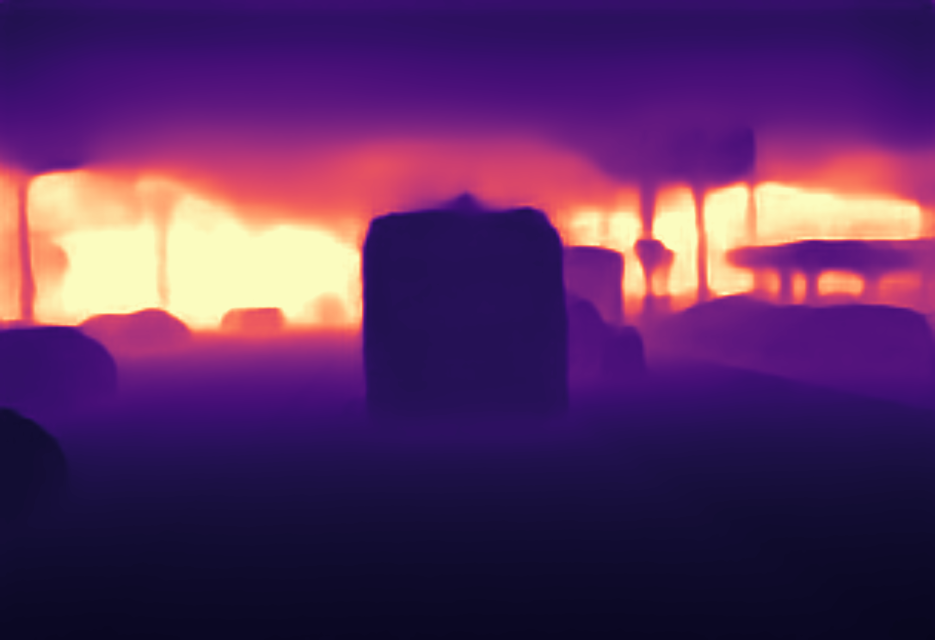} & \includegraphics[height=\turnheightnew]{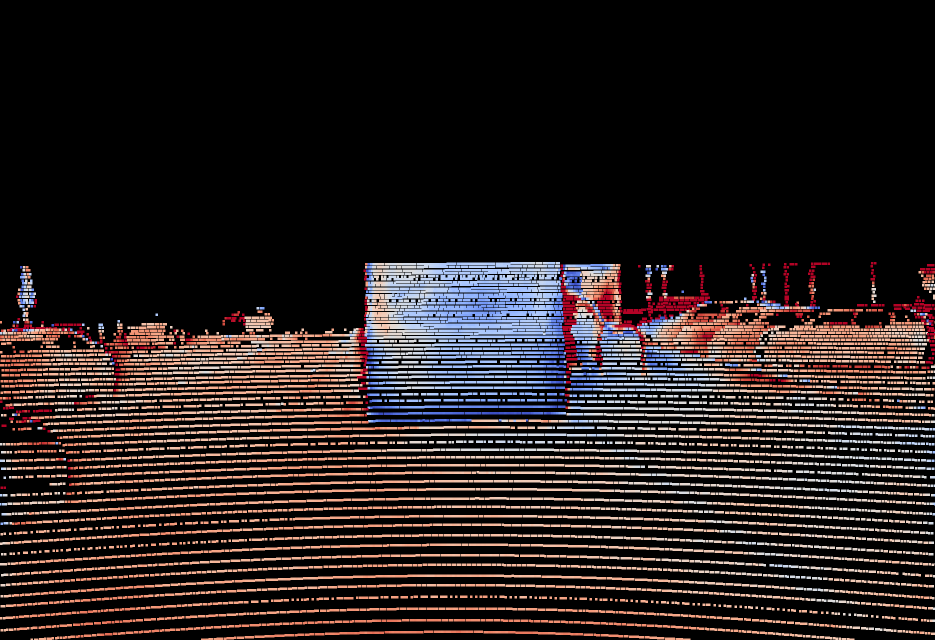} &
		\includegraphics[height=\turnheightnew]{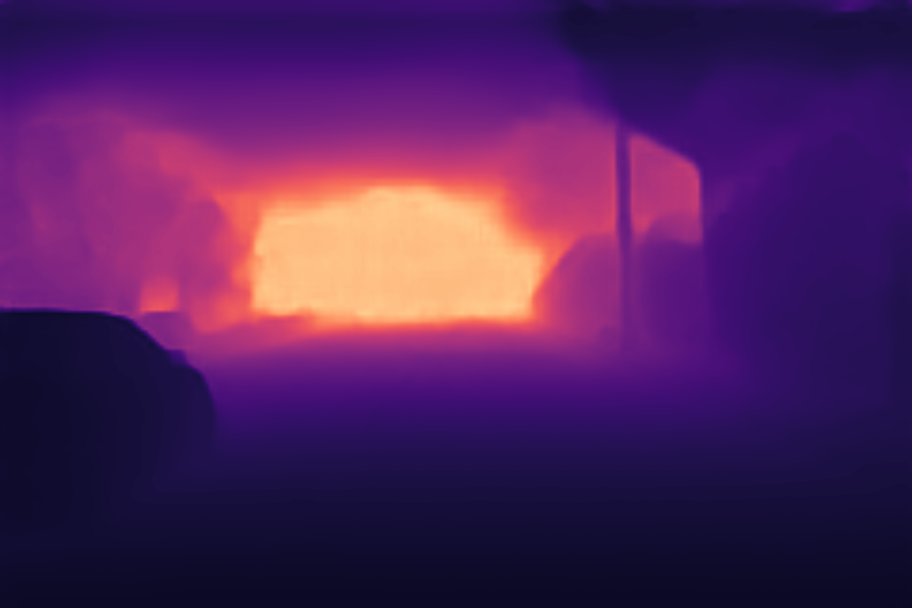} & \includegraphics[height=\turnheightnew]{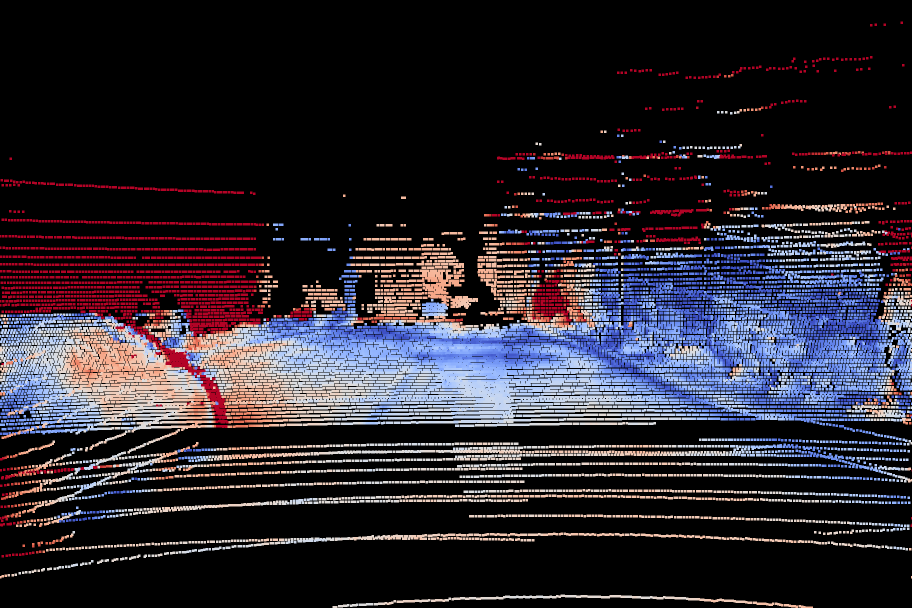}&\\
		
		\raisebox{0.5\normalbaselineskip}[0pt][0pt]{\rotatebox{90}{Vertical}}& 
		\includegraphics[height=\turnheightnew, trim=120pt 0 120pt 0, clip]{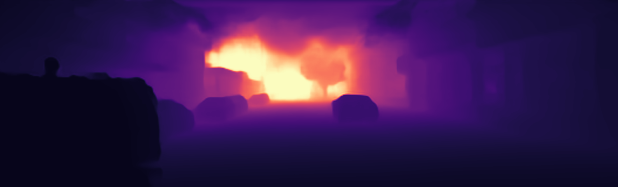} & \includegraphics[height=\turnheightnew, trim=120pt 0 120pt 0, clip]{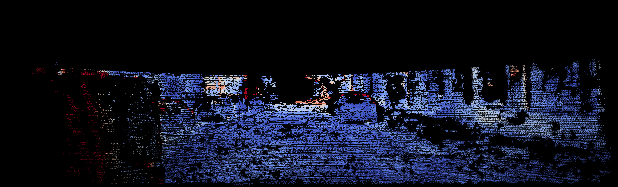} &
		\includegraphics[height=\turnheightnew]{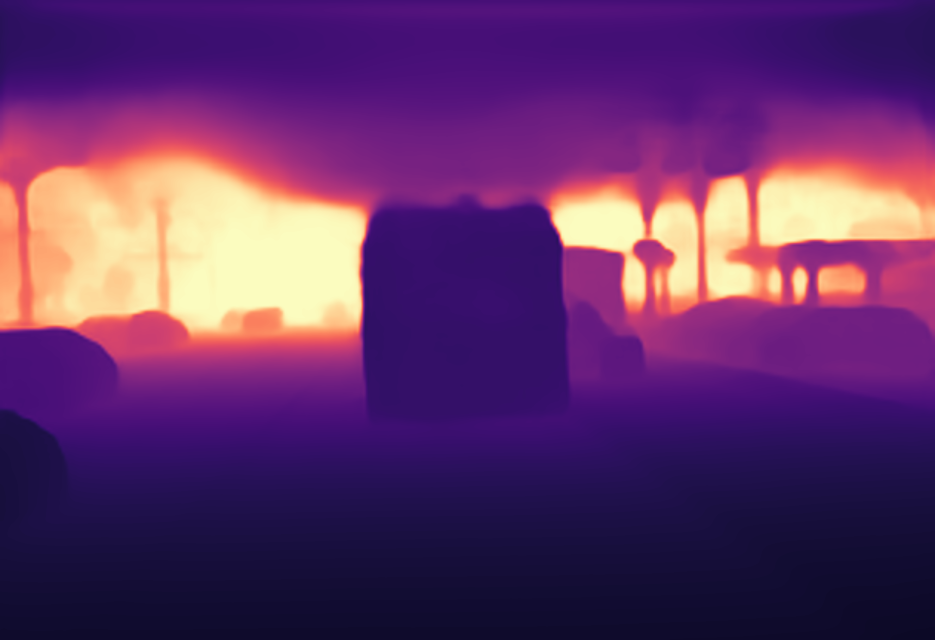} & \includegraphics[height=\turnheightnew]{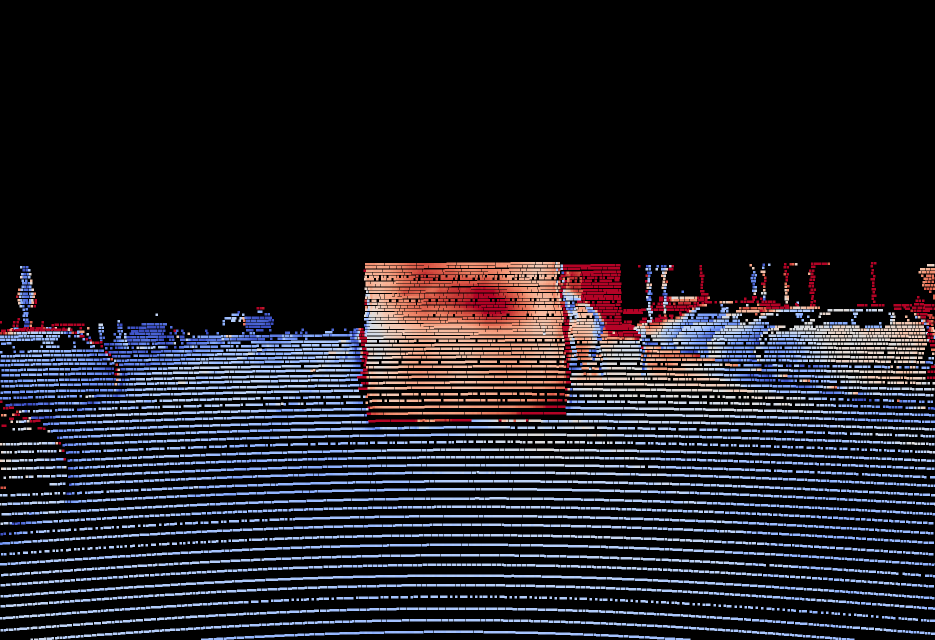} &
		\includegraphics[height=\turnheightnew]{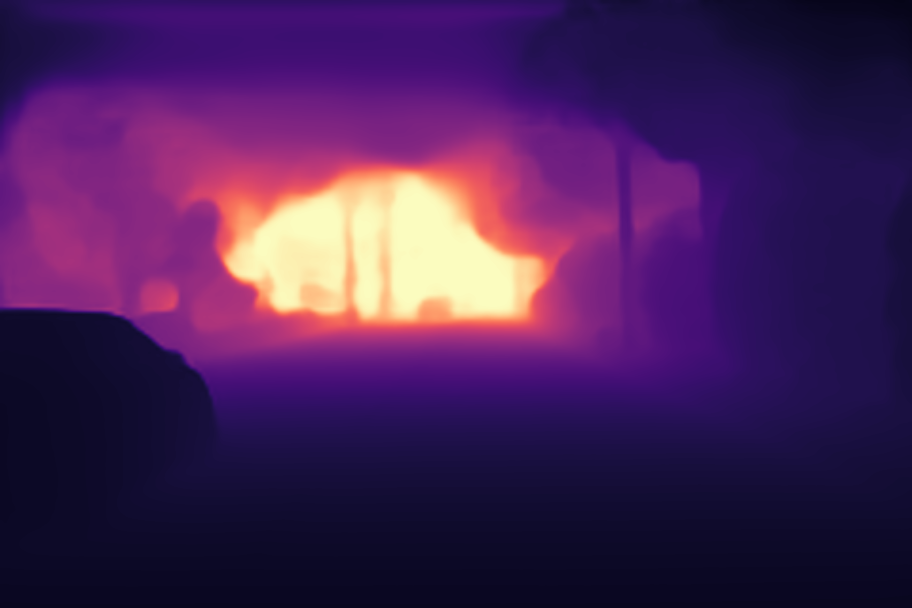} & \includegraphics[height=\turnheightnew]{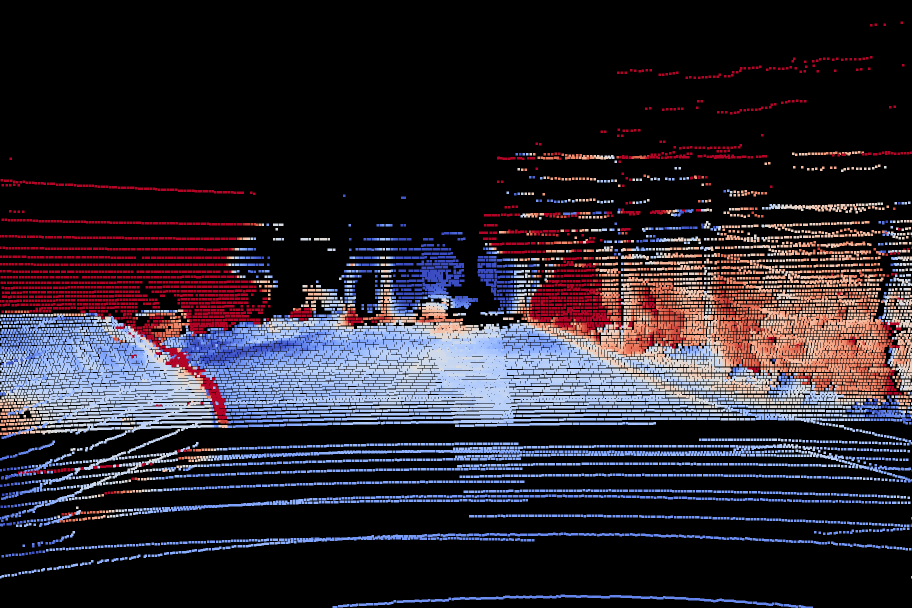}&\\
		
		\raisebox{0.7\normalbaselineskip}[0pt][0pt]{\rotatebox{90}{Fusion}}& 
		\includegraphics[height=\turnheightnew, trim=120pt 0 120pt 0, clip]{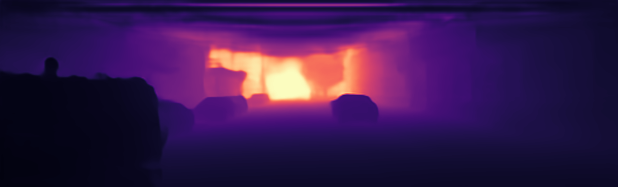} & \includegraphics[height=\turnheightnew, trim=120pt 0 120pt 0, clip]{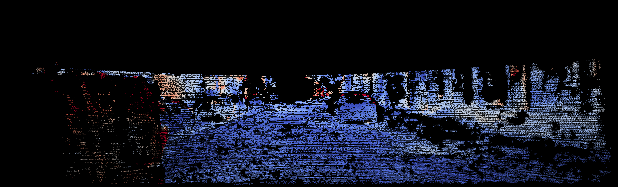} &
		\includegraphics[height=\turnheightnew]{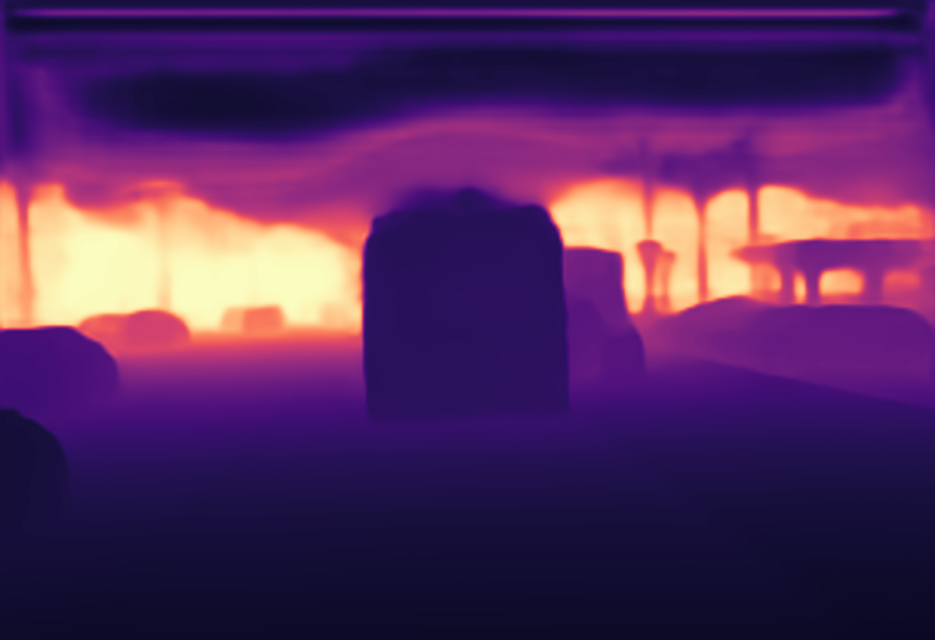} & \includegraphics[height=\turnheightnew]{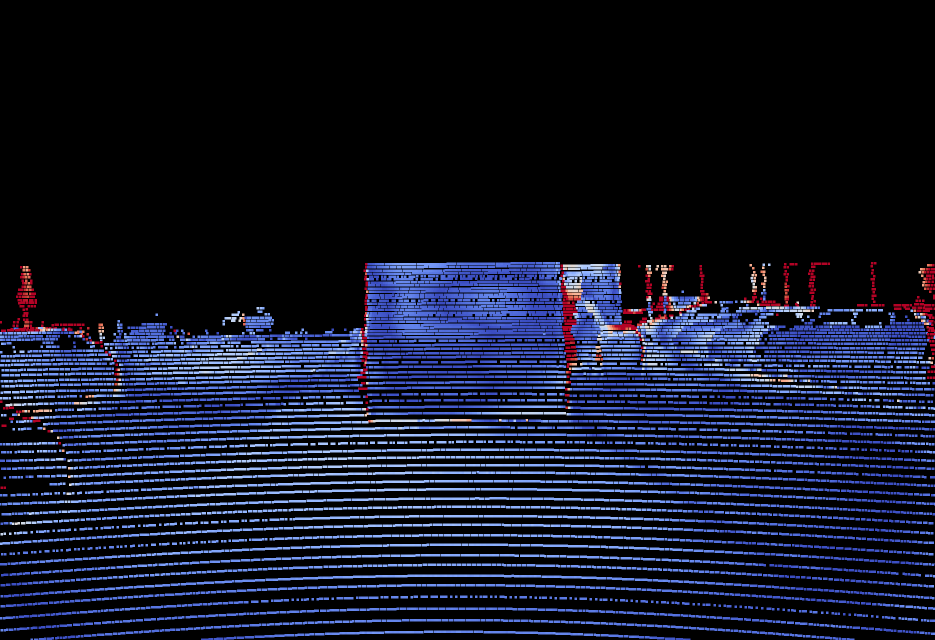} &
		\includegraphics[height=\turnheightnew]{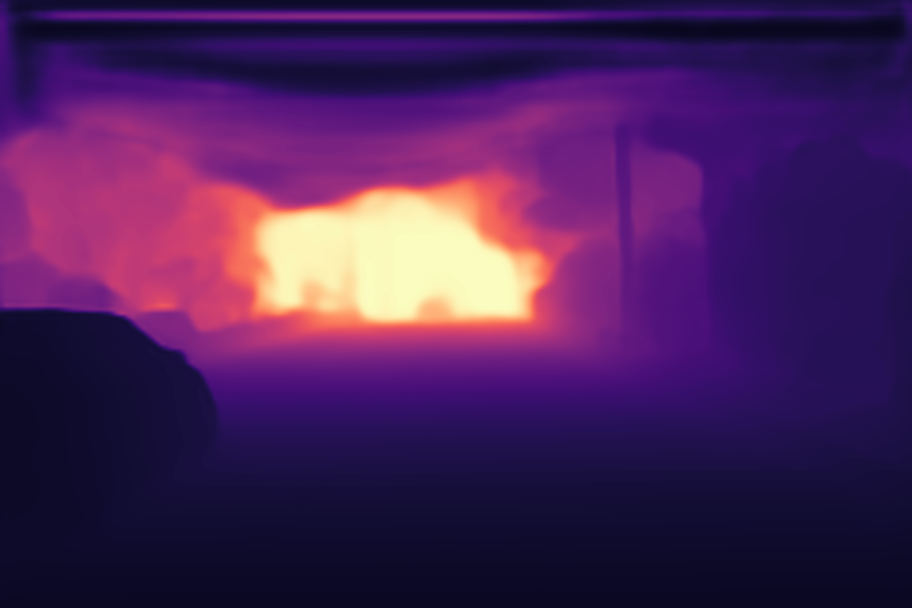} & \includegraphics[height=\turnheightnew]{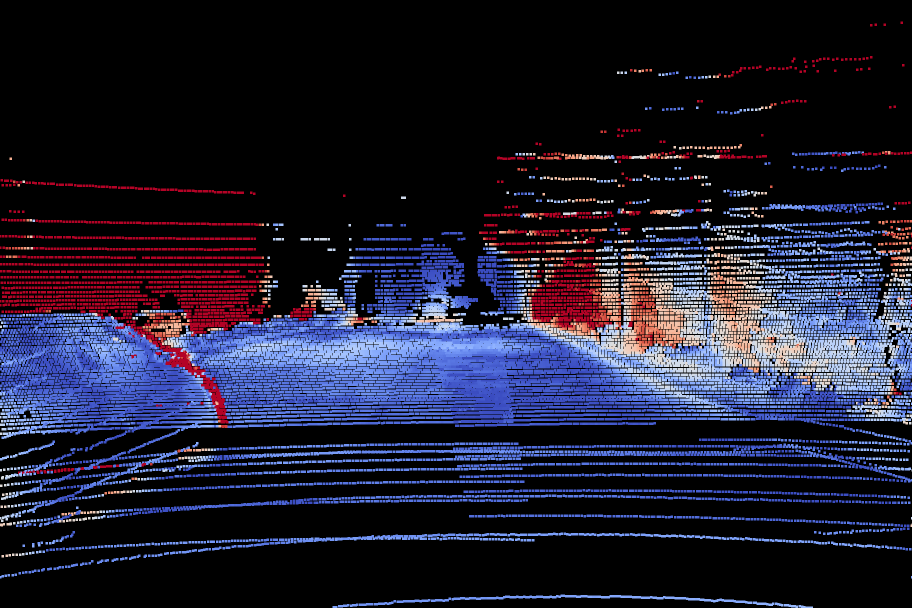}&\\

	\end{tabular}
	\captionof{figure}{\textbf{Qualitative ablation.} Predicted depths and errors for DrivingStereo $\rightarrow$ \{KITI, Waymo, DDAD\} models. Baseline -- standard depth regression with geometric augmentations. Focal -- depth regression via Focal Canonical Transform - $\mathrm{FCT}_{\bm{\mathcal{C}}}(\cdot)$. This is equivalent to Metric3D\cite{yin2023metric3d}, but trained on our setup. Vertical -- depth regression via Vertical Canonical Transform - $\mathrm{VCT}_{\bm{\mathcal{C}}}(\cdot)$. Fusion -- depth regression with uncertainty-based fusion model.}
	\label{fig:ablation}

\end{center}
}]

\section{Qualitative Results}
Qualitative results for different model configurations are shown in \cref{fig:ablation}. All models demonstrate comparable capabilities in reconstructing semantic objects, effectively capturing object boundaries and overall scene layout. However, error maps reveal that the Baseline model struggles to handle the domain gap introduced by varying camera parameters in the zero-shot transfer setting, leading to reduced accuracy. Both the Focal and Vertical models perform reasonably well across most image regions, though they exhibit noticeable modeling errors in certain areas. In contrast, the Fusion model attains the highest accuracy by adaptively integrating the depth predictions from both Focal and Vertical models.

\section{Additional Ablations with Ground Plane Methods}
\noindent{}\textbf{Additional Results.}
\Cref{tab:gedepth} presents a more detailed ablation study of GEDepth \cite{yang2023gedepth}, incorporating results from additional training datasets. GVDepth consistently outperforms GEDepth across multiple training-test dataset combinations.
We do not report results on Argoverse and DDAD datasets, as GEDepth failed to achieve satisfactory convergence. We will provide further comments on this issue below. 

Furthermore, in \cref{tab:ground}, we provide results for PlaneDepth \cite{wang2023planedepth} and GroCo \cite{cecille2024groco}, which utilize the ground plane constraint in a similar manner as GVDepth. Since these methods are self-supervised, we retrained PlaneDepth with supervision, using the same backbone as our method. The results suggest that the performance gap is not due to difference in supervision or model complexity, but due to the superior out-of-distribution generalization of our approach. For GroCo \cite{cecille2024groco}, we report the results from the original paper, as it lacks the open-source code implementation. However, GroCo is fundamentally a self-supervised version of GEDepth \cite{yang2023gedepth}, preserving all of its core principles. 

Our main argument is that none of these methods are explicitly designed to enhance generalization or facilitate training with diverse perspective geometries. Moreover, they improve performance only for road pixels, unlike the proposed $\mathrm{VCT}_{\bm{\mathcal{C}}}(\cdot)$ that exploits the ground plane constraint for all objects.
In the following sections, we provide an in-depth analysis, offering insights into the generalization and convergence challenges faced by this methods.
\begin{table}[t]
	\centering
	\caption{\textbf{Comparison with GEDepth \cite{yang2023gedepth}.} All models are trained with equivalent setup and model complexity on KITTI, DrivingStereo and Waymo datasets. Best results are \textbf{bolded}, second best are \underline{underlined}. In-domain evaluation results are \colorbox{lightgray}{shaded}.}
	\vspace{-5pt}
	\resizebox{1.0\linewidth}{!}{
		\begin{tabular}{cc|cc|cc|cc}
			\toprule[0.4mm]
			& \textbf{Testing} & \multicolumn{2}{c|}{\textbf{GEDepth}\cite{yang2023gedepth}}  & \multicolumn{2}{c|}{\textbf{Vertical}} & \multicolumn{2}{c}{\textbf{Fusion}} \\
			
			& \textbf{dataset} & $\mathrm{A.Rel}\downarrow$ & $\mathrm{\delta}_{1}\uparrow$ & $\mathrm{A.Rel}\downarrow$ & $\mathrm{\delta}_{1}\uparrow$ & $\mathrm{A.Rel}\downarrow$ & $\mathrm{\delta}_{1}\uparrow$ \\
			
			\midrule
			
				\multirow{5}{*}{\begin{turn}{90}\textbf{KITTI}\end{turn}}& KITTI & \cellcolor{lightgray}5.68 & \cellcolor{lightgray}95.4 & \cellcolor{lightgray}5.70 &\cellcolor{lightgray}95.5 &\cellcolor{lightgray}\textbf{5.67} & \cellcolor{lightgray}\textbf{95.7}\\ 
			& DStereo &18.25 &66.7 & \underline{10.43} & \underline{87.3} & \textbf{10.24} & \textbf{87.4} \\
			& Waymo & 22.11 & 69.1 & \textbf{13.42} & \textbf{78.6} & \underline{14.34} & \underline{78.4}\\
			& Argo & 19.27 & 52.8 & \underline{14.72} & \underline{64.7} & \textbf{10.45} &\textbf{84.8} \\
			& DDAD & 19.46 & 61.2 & \underline{14.26} & \underline{73.6} & \textbf{11.81} & \textbf{82.5} \\
			\midrule
			\multirow{5}{*}{\begin{turn}{90}\textbf{Dstereo}\end{turn}}& KITTI & 11.77 & 74.2 & \underline{7.42} & \underline{92.6} & \textbf{6.96} & \textbf{92.7}\\
			& DStereo & \cellcolor{lightgray}\textbf{2.99} &\cellcolor{lightgray}\textbf{99.5} & \cellcolor{lightgray}3.07 &\cellcolor{lightgray}99.5& \cellcolor{lightgray}\underline{3.01} &\cellcolor{lightgray}\underline{99.5}\\
			& Waymo & 18.88 & 73.3 & \textbf{11.72} & \textbf{85.5} & \underline{12.15} & \underline{83.1}\\
			& Argo & 13.19 & 83.0 & \underline{11.41} & \underline{84.7} & \textbf{9.91} & \textbf{86.9} \\
			& DDAD & 20.20 & 55.18 & 15.71 & 80.0 & \textbf{12.02} & \textbf{82.4}\\
			
			\midrule
			\multirow{5}{*}{\begin{turn}{90}\textbf{Waymo}\end{turn}}& KITTI & 15.91  & 82.1 & \textbf{8.30} & \textbf{92.3} & \underline{10.92} & \underline{89.8}\\
			& DStereo & 14.21 & 84.4 & \textbf{11.40} & \underline{86.8} & \underline{12.78} & \textbf{87.1}\\
			& Waymo & \cellcolor{lightgray}\textbf{3.42} &\cellcolor{lightgray}\textbf{99.0}  &\cellcolor{lightgray}3.61 &\cellcolor{lightgray}98.9 &\cellcolor{lightgray}\underline{3.51} &\cellcolor{lightgray}\underline{98.8} \\
			& Argo & 17.83 & 83.4 & \underline{12.21} & \underline{88.6} & \textbf{9.62} & \textbf{94.1} \\
			& DDAD & 19.20 & 66.6 & \textbf{11.51} & \underline{84.4} & \underline{13.99} &\textbf{86.2} \\

			\bottomrule[0.4mm]
	\end{tabular}}
\label{tab:gedepth}

\end{table}
\begin{table}[t]
	\centering
	\caption{\textbf{Comparison with self-supervised methods that leverage the known ground plane.} Results for KITTI $\rightarrow$ DDAD zero-shot transfer.(\dag): Trained with supervision. (\ddag): Results taken from corresponding paper.}
	\resizebox{0.9\linewidth}{!}{
		\begin{tabular}{l|cccc}
			\toprule[0.4mm]
			
			\textbf{Method}& $\mathrm{A.Rel}\downarrow$  & $\mathrm{RMS}\downarrow$ & $\mathrm{RMS}_{\log}\downarrow$ & $\mathrm{\delta}_{1}\uparrow$\\
			
			\midrule
			PlaneDepth\textsuperscript{\dag}~\cite{wang2023planedepth} & 30.28 & 12.01 & 0.499 & 37.8 \\
			PlaneDepth~\cite{wang2023planedepth} & 32.71 & 14.77 & 0.529 & 37.9 \\
			GroCo\textsuperscript{\ddag} ~\cite{cecille2024groco} & 42.40 & 15.37 & - & -\\
			
			Vertical\dag & \underline{14.26} & \underline{9.39} & \underline{0.271} & \underline{73.6}\\ 
			Fusion\dag & \textbf{11.81} & \textbf{8.35} & \textbf{0.251} & \textbf{82.5}\\ 
			
			\bottomrule[0.4mm]
	\end{tabular}}
\label{tab:ground}

\end{table}

\noindent{}\textbf{Detailed Analysis.}
\begin{figure}
	\includegraphics[width=\linewidth]{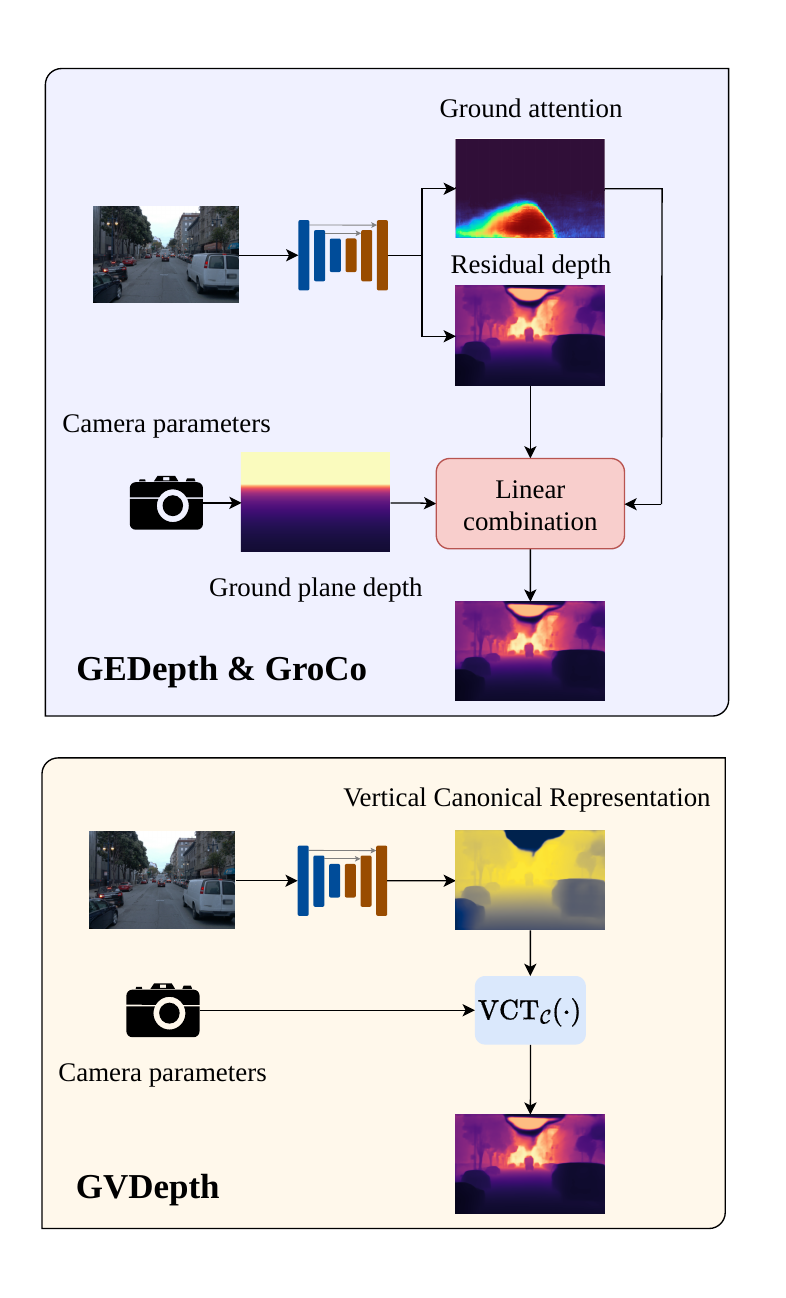}
	\caption{\textbf{Different approaches for incorporation of ground plane constraint.} 
		In our approach, $\mathrm{VCT}_{\bm{\mathcal{C}}}(\cdot)$ serves as a \textit{point of disentanglement}, which enables learning of \textit{Vertical Canonical Representation} that is invariant to specific perspective geometry. Linear combination used in GEDepth and GroCo does not hold such properties, leading to limited generalization.}
	\label{fig:gedepth}
\end{figure}
We start with the following premises, which differentiate the proposed $\mathrm{VCT}_{\bm{\mathcal{C}}}(\cdot)$ from aforementioned approaches:
\begin{enumerate}[label=(\roman*)]
	\item Unlike GEDepth and GroCo, which utilize the ground plane constraint only for road pixels, the proposed canonical representation enables utilization of vertical image position cue for all objects in the scene;
	\item Similarly, in constrast to GEDepth and GroCo, our approach is inherently modeled to enable perspective geometry disentanglement for all objects, resulting in better out-of-distribution generalization.
\end{enumerate}
While our extensive evaluation confirms this, here we also provide insights into why GEDepth and GroCo fail to generalize across all objects in the scene.

Both GEDepth and GroCo formulate the problem of depth regression $\mathbf{D} \in [0, D_{max}]^{H \times W}$ as a linear combination of the known ground depth $\mathbf{D_G} \in [0, D_{max}]^{H \times W}$ and the learnable residual depth $\mathbf{D_R} \in [0, D_{max}]^{H \times W}$, weighted by the learnable \textit{``ground attention"} $\mathbf{A} \in [0, 1]^{H \times W}$:
\begin{equation}
	\mathbf{D} = \mathbf{A} \odot \mathbf{D_G} + (1 - \mathbf{A}) \odot \mathbf{D_R}.
	\label{eqn:gedepth}
\end{equation}
Through a straightforward analysis, we can notice why this formulation enhances depth estimation for road pixels only.
First, any part of the scene appearing above the horizon can not benefit from the induced perspective geometry constraint, as $\mathbf{D_G}$ is invalid for those regions of the image.
Moreover, ground attention $\mathbf{A}$ is usually estimated with the $\mathrm{Sigmoid}(\cdot)$, which has a natural tendency to converge to either 0 or 1 during optimization, leading to $\mathbf{A}$ effectively being equivalent to road segmentation, as visualized in \cite{cecille2024groco, yang2023gedepth}.
Even for rare cases where this is not the case, the resulting linear combination has an ambivalent geometric interpretation chosen arbitrarily by the model.
Our proposed $\mathrm{VCT}_{\bm{\mathcal{C}}}(\cdot)$ is a simple and elegant solution for all of these issues; it enables effective utilization of the ground plane constraint for all image regions while preserving a clear geometric interpretation.

The same fundamental principles can be applied to perspective geometry disentanglement. While $\mathbf{D_G}$ clearly disentangles depth from camera parameters, the estimated $\mathbf{D_R}$ does not hold such properties, inducing generalization errors due to the ambiguity of depth and camera parameters.
Since the final depth $\mathbf{D}$ is calculated as a linear combination, it is only fully disentangled where $\mathbf{A}$ is 1, which is valid only for road pixels.
On the other hand, $\mathrm{VCT}_{\bm{\mathcal{C}}}(\cdot)$ is inherently designed to incorporate perspective geometry disentanglement, regardless of the specific image region.

\noindent{}\textbf{GEDepth Convergence Issues.}
GEDepth optimizes $\mathbf{D}$ from \cref{eqn:gedepth} without additional regularization factors or external segmentation modules, claiming that the learned attention map can automatically separate ground and other regions.
Unfortunately, in our experiments this occurred rather inconsistently, and the learned attention map converged to adequate ground segmentation only on the 19th training experiment.
In most cases, the model resorted to estimating $\mathbf{A} = \mathbf{0}^{H \times W}$, thus ignoring the provided ground plane information and resulting in $\mathbf{D} = \mathbf{D_R}$.

\noindent{}\textbf{Final Remarks.}
We conducted a thorough evaluation that highlights the extremely limited generalization capabilities of models that rely on ground plane constraints \cite{wang2023planedepth, yang2023gedepth, cecille2024groco}. While one might assume that these issues arise from limited reproducibility or incorrect convergence, the analysis we presented in previous sections suggests that the root cause lies in inherent design choices. These models are optimized to perform well only on narrow training distribution. Even within these constraints, improvements in accuracy are confined to road pixels, which are of limited relevance for safety-critical applications. In contrast, the proposed $\mathrm{VCT}_{\bm{\mathcal{C}}}(\cdot)$ elegantly addresses all these limitations.

\section{Model Architecture}

In this work, we employ a fully convolutional encoder-decoder architecture with skip connections.
While vision transformer (ViT)-based encoders, such as DINOv2 \cite{oquab2023dinov2}, often achieve superior accuracy, their advantages typically rely on large-scale training.
A similar fully convolutional design is adopted in Metric3D \cite{yin2023metric3d}, demonstrating the effectiveness of such models for generalizable monocular depth estimation (MDE).
Given our focus on single-dataset training, we prioritize convolutional backbones to reduce computational complexity while maintaining robust generalization performance.

\noindent{}\textbf{Decoder Architecture Details.}
Our decoder is designed similarly as in Metric3D \cite{yin2023metric3d}, consisting of four blocks that progressively upsample and fuse encoder features from a resolution of $(\frac{H}{32}, \frac{W}{32})$ to $(\frac{H}{2}, \frac{W}{2})$. 
Decoder channels dimensions are $\{756, 512, 256, 128\}$.
Our fusion module processes $(\frac{H}{2}, \frac{W}{2})$ resolution feature maps $\mathbf{F}$ with two compact U-Net-like networks \cite{ronneberger2015u}, each containing 3 downsampling and upsampling blocks, producing two final feature representations.
These feature representations are processed with two convolutional blocks which predict our canonical representations $(\mathbf{C_F}, \mathbf{C_Y})$ and accompanying uncertainty estimates $(\mathbf{\Sigma_F}, \mathbf{\Sigma_Y})$.
After uncertainty-based fusion, final depth map is bilinearly upsampled to $(H, W)$ resolution.

\begin{figure}
	\includegraphics[width=\linewidth, trim=20pt 0pt 25pt 10pt, clip]{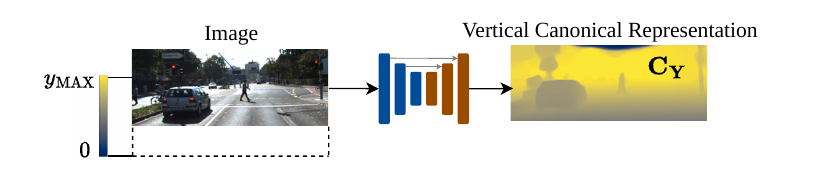}
	\caption{\textbf{Vertical Canonical Representation regression.} 
		Visualization of regression boundaries of proposed \textit{Vertical Canonical Representation.} We predict vertical image position of ground plane projection in range $[0, y_{MAX}]$. Image is \textit{\textbf{virtually extended}} beyond bottom boundary to enable depth estimation for pixels with ground-plane projection below camera field-of-view. For stability reasons, regression is bounded to $y_{MAX}$, which corresponds to $d_{MAX}=80$.}
	\label{fig:vertical_regression}
\end{figure}

\noindent{}\textbf{Vertical Canonical Representation details.}
For convenience, we restate the equation used in Vertical Canonical Transform $\text{VCT}_{\bm{\mathcal{C}}}(\cdot)$: 
\begin{equation}
	d=\frac{f_y h}{\left(H-c_y-y\right) \cos (\theta)-f_y \sin (\theta)}.
	\label{eqn:vertical_supp}
\end{equation}
Upon closer inspection, it becomes evident that mapping $y \mapsto d$ can lead to unstable training if not properly regularized. Specifically,
at $y$ corresponding to the horizon level of a given camera setup, the depth $d$ approaches infinity.
Furthermore, for any $y$ above the horizon, the depth $d$ becomes negative.
To address this, as visualized in \cref{fig:vertical_regression}, we bound the regression of $y$ to $y_{MAX}$, derived from the inverse mapping from $d_{MAX}=80$.
Additionally, to accommodate depth regression for objects with ground-plane projections below the camera’s field of view, we \textit{virtually extend} the image. In practice, this corresponds to increasing $H$ in \cref{eqn:vertical_supp} accordingly.

\section{Model Complexity}
\begin{table}[!h]
	\centering
	\caption{\textbf{Model architecture details.} Encoder architectures number of parameters and execution time for models used in this work. Execution time is measured for resolution 416x640, on a single NVIDIA RTX A6000 GPU.}
	\resizebox{\columnwidth}{!}{
\begin{tabular}{ccccc}
	\toprule[0.4mm]
		Method & Encoder & Params \# & Execution time
		\\
		\midrule
		Monodepth2 \cite{godard2019digging} & ResNet-50 \cite{he2016deep} & 34M & 12.70 ms \\
		DIFFNet \cite{zhou_diffnet} & HRNet-18 \cite{wang2020deep} & 11M & 29.42 ms \\
		NeWCRFs \cite{yuan2022newcrfs} & Swin-L \cite{liu2021swin} & 270M & 62.56 ms \\
		iDisc \cite{piccinelli2023idisc} & Swin-L \cite{liu2021swin} & 208M & 120.51 ms \\
		
		PlaneDepth \cite{wang2023planedepth} & ResNet-50 \cite{he2016deep} & 39M & 15.21 ms \\
		\midrule
		MiDaS \cite{ranftl2020towards} & ResNeXt-101 \cite{xie2017aggregated} & 105M & 26.51 ms \\
		LeReS \cite{yin2021learning} & ResNet-50 \cite{he2016deep} & 52M & 20.26 ms \\
		\midrule
		ZeroDepth \cite{guizilini2023towards} & ResNet-18 \cite{he2016deep} & 232M & 238.48 ms \\
		Metric3D \cite{yin2023metric3d} & ConvNext-L \cite{liu2022convnet} & 203M & 10.74 ms \\ 
		UniDepth \cite{piccinelli2024unidepth} & ConvNext-L \cite{liu2022convnet} & 239M & 55.34 ms \\
		\midrule
		\textbf{GVDepth} & ConvNext-L \cite{liu2022convnet} & 228M & 21.24 ms \\
		\bottomrule[0.4mm]
	\end{tabular}}
	\label{tbl:complexity}
\end{table}
In \cref{tbl:complexity} we present the architectural details of the models used in this work. 
While most results align with expectations, there are a few notable outliers worth discussing.
First, ZeroDepth exhibits the highest inference times, despite utilizing the least complex backbone among all methods.
Essentially, ZeroDepth shifts complexity from the pre-trained backbone to its proprietary self-attention layers in the decoder.
This trade-off limits its ability to fully leverage the benefits of large-scale pretraining, which may explain its subpar accuracy compared to UniDepth, Metric3D, and GVDepth, even though it uses significantly more data during training.

On the other hand, Metric3D achieves remarkably low latency in depth prediction despite employing a relatively complex backbone.
This efficiency likely stems from predicting depths at a reduced $(\frac{H}{4}, \frac{W}{4})$ resolution, which is subsequently upsampled to the original resolution
Lastly, GVDepth stands out for its lightweight design and low inference time, even though it uses a relatively complex backbone.
Its impressive generalization performance is attributed to our novel methodology rather than reliance on model or data scaling.
In future work, we will investigate more complex transformer architectures, which were costly to train on our current computational setup.

\section{Ablation of Fusion Strategy.}
\begin{table}[t]
	\centering
	\caption{\textbf{Ablation of fusion strategies.} Mean -- fusion via mean of $\mathrm{FCT}_{\bm{\mathcal{C}}}(\cdot)$ and $\mathrm{VCT}_{\bm{\mathcal{C}}}(\cdot)$. L1 -- Uncertainty loss without aleatoric uncertainty weighting in $\mathcal{L}_{unc}$. All models are trained on DrivingStereo.}
	\resizebox{\linewidth}{!}{
		\begin{tabular}{c|cc|cc|cc}
			\toprule[0.4mm]
			
			\multirow{2}{*}{\textbf{Configuration}} &
			\multicolumn{2}{c|}{\textbf{DrivingStereo}}&  \multicolumn{2}{c|}{\textbf{KITTI}}  & \multicolumn{2}{c}{\textbf{Waymo}}\\
			
			& $\mathrm{A.Rel}\downarrow$  & $\mathrm{\delta}_{1}\uparrow$ & $\mathrm{A.Rel}\downarrow$  & $\mathrm{\delta}_{1}\uparrow$ & $\mathrm{A.Rel}\downarrow$ & $\mathrm{\delta}_{1}\uparrow$ \\
			
			\midrule
			Mean &\cellcolor{lightgray}\textbf{3.05}&\cellcolor{lightgray}\underline{99.4}& 8.21 & 91.1 & 14.01 & 82.8 \\
			L1 &\cellcolor{lightgray}3.09&\cellcolor{lightgray}99.4& \underline{8.02} & \underline{91.4} & \underline{13.23} & \underline{82.9} \\
			Fusion &\cellcolor{lightgray}\underline{3.07}&\cellcolor{lightgray}\textbf{99.5}& \textbf{6.96} & \textbf{92.7} & \textbf{12.15} & \textbf{83.1} \\ 
			
			\bottomrule[0.4mm]
	
	\end{tabular}}
	\label{tab:uncertainty}
\end{table}
In \cref{tab:uncertainty} we examine the efficacy of different fusion strategies.
Zero-shot transfer results demonstrate that our adaptive fusion weighted by aleatoric uncertainties leads to superior generalization performance.

\section{Dataset Details}
In this work, we use KITTI \cite{geiger2013vision}, DDAD \cite{guizilini20203d}, DrivingStereo \cite{yang2019drivingstereo}, Waymo \cite{Sun_2020_CVPR} and Argoverse Stereo \cite{Argoverse} datasets, both for training and evaluation.
For KITTI dataset, we evaluate all models on commonly used Eigen split \cite{eigen2014depth} with Garg crop \cite{garg2016unsupervised}, resulting in 23158 training images and 652 testing images.
On DDAD dataset we use the official training and validation split, with 12650 and 3950 images, respectively.
Since Waymo, DrivingStereo, and Argoverse Stereo are not widely used for MDE evaluation, we simplify the process by creating custom dataset splits. The resulting training splits consist of \{156K, 168K, 5K\} samples, while the corresponding testing splits contain \{5K, 5K, 500\} samples, respectively.

\section{Camera setup calibration}
\label{supp:calib}

In this section, we provide additional details about our camera calibration procedure.
Our proposed Vertical Canonical Transform $\text{VCT}_{\bm{\mathcal{C}}}(\cdot)$, as indicated in 
\cref{eqn:vertical_supp}, requires the knowledge of camera parameters $\bm{\mathcal{C}} = \{f_y, c_y, h, \theta\}$.
Here, for all datasets, $f_y$ and $c_y$ are usually known up to the reasonable error induced by the calibration procedure.
However, for certain datasets, camera height $h$ and camera pitch $\theta$ are either unknown, or not properly calibrated.
To remain consistent throughout this work, we recalibrate the extrinsic parameters for each dataset, with details provided in \cref{alg:calib}.
\begin{algorithm}[t]
	\caption{Estimate Camera Height $h$ and Pitch $\theta$.}
	\begin{algorithmic}[1]
		\Require RGB images $\{I_i\}_{i=1}^N$, ground-truth depth maps $\{D_i\}_{i=1}^N$
		\Ensure Median camera height $h_{\text{median}}$ and pitch $\theta_{\text{median}}$
		\State Initialize empty sets $\mathcal{H} = \emptyset$, $\Theta = \emptyset$ 
		\For{$i = 1$ to $N$} 
		\State Run semantic segmentation model on $I_i$ to acquire road pixels $\mathcal{P}_i$ 
		\[
		\mathcal{P}_i = \{(u, v) \mid (u, v) \text{ belongs to the road in } I_i\}
		\]
		\State Extract depths $\{D_i(u, v) \mid (u, v) \in \mathcal{P}_i\}$
		\State Filter road pixels based on depth:
		\[
		\mathcal{P}_i^{\text{filtered}} = \{(u, v) \in \mathcal{P}_i \mid D_i(u, v) < 20\}
		\]
		\State Project filtered road pixels to 3D points:
		\[
		\mathcal{Q}_i = \left\{ \begin{pmatrix} X \\ Y \\ Z \end{pmatrix} \Bigg| 
		\begin{aligned}
			X &= \frac{(u - c_x) Z}{f_x}, \\
			Y &= \frac{(v - c_y) Z}{f_y}, \\
			Z &= D_i(u, v)
		\end{aligned}, \forall (u, v) \in \mathcal{P}_i^{\text{filtered}} \right\}
		\]
		\State Fit $\mathbf{R}_i(\theta_i)$ (rotation matrix) and $h_i$ (camera height) to $\mathcal{Q}_i$ with RANSAC plane estimation
		\State Append $h_i$ to $\mathcal{H}$ and $\theta_i$ to $\Theta$
		\EndFor
		\State Compute median camera height and pitch:
		\[
		h_{\text{median}} = \text{median}(\mathcal{H}), \quad \theta_{\text{median}} = \text{median}(\Theta)
		\]
		\State \Return $h_{\text{median}}, \theta_{\text{median}}$
	\end{algorithmic}
\label{alg:calib}
\end{algorithm}
For semantic segmentation of the road plane we use the DeepLabv3 model \cite{chen2018encoder}.

\begin{table}[!t]
	\centering
	\caption{\textbf{Camera calibration}. Camera extrinsics estimated by the calibration procedure described in \cref{alg:calib}.}
	\label{tbl:datasets}

		\begin{tabular}{c|cc}
			\toprule[0.5mm]
			
			Dataset & $h$[m]  & $\theta$[\textdegree] \\
			\midrule
			Argoverse Stereo \cite{Chang_2019_CVPR} & 1.678 & 0.021 \\
			DDAD \cite{guizilini20203d} & 1.459 & -0.519 \\
			DrivingStereo \cite{yang2019drivingstereo} & 1.739 & -0.561 \\
			KITTI \cite{geiger2013vision} & 1.659 & -0.664 \\
			Waymo \cite{Sun_2020_CVPR} & 2.145 & -0.331 \\
			
			\bottomrule[0.5mm]
	\end{tabular}
	\label{tbl:calib}
\end{table}

\begin{figure}[t]
	\newcommand{\turnwidthnew}{0.47\columnwidth}
	\centering
	\small
	\begin{tabular}{@{\hskip 1mm}r@{\hskip 1mm}c@{\hskip 1mm}c@{\hskip 0mm}}
		
		& \quad Camera height & \quad Camera pitch \\
		
		\raisebox{1.8\normalbaselineskip}[0pt][0pt]{\rotatebox{90}{Argoverse \cite{Chang_2019_CVPR}}}& 
		\includegraphics[width=\turnwidthnew, trim=0pt 0 40pt 40pt, clip]{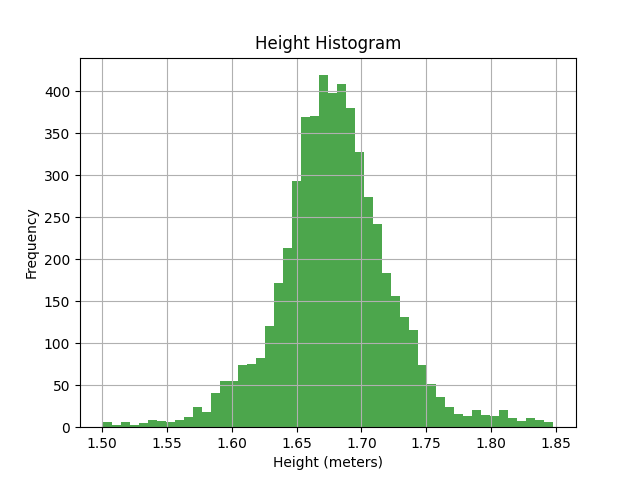} & \includegraphics[width=\turnwidthnew, trim=0pt 0 40pt 40pt, clip]{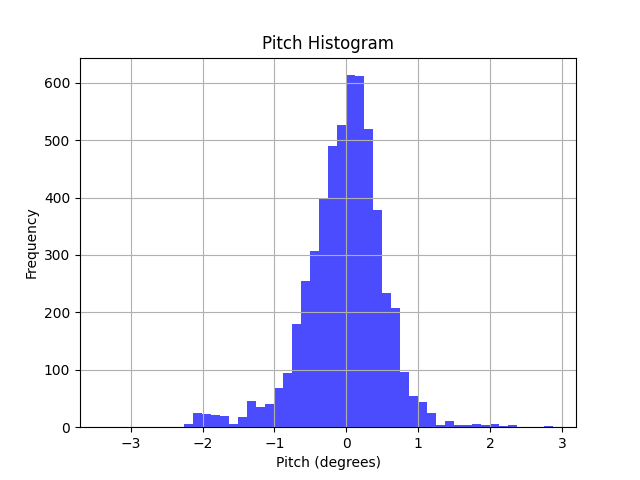}  \\
		
		\raisebox{2.2\normalbaselineskip}[0pt][0pt]{\rotatebox{90}{DDAD \cite{guizilini20203d}}}& 
		\includegraphics[width=\turnwidthnew, trim=0pt 0 40pt 40pt, clip]{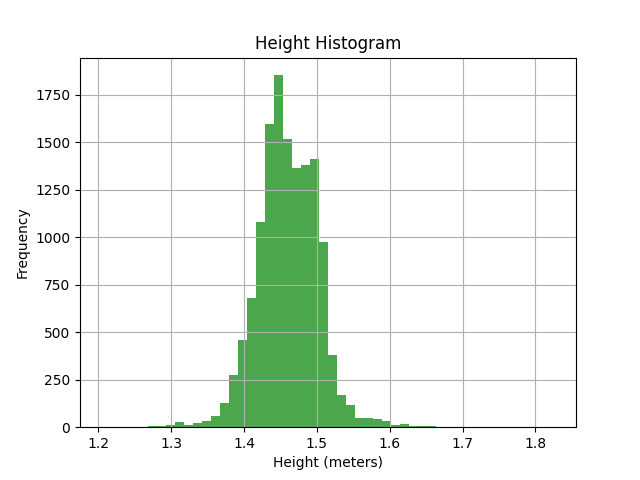} & \includegraphics[width=\turnwidthnew, trim=0pt 0 40pt 40pt, clip]{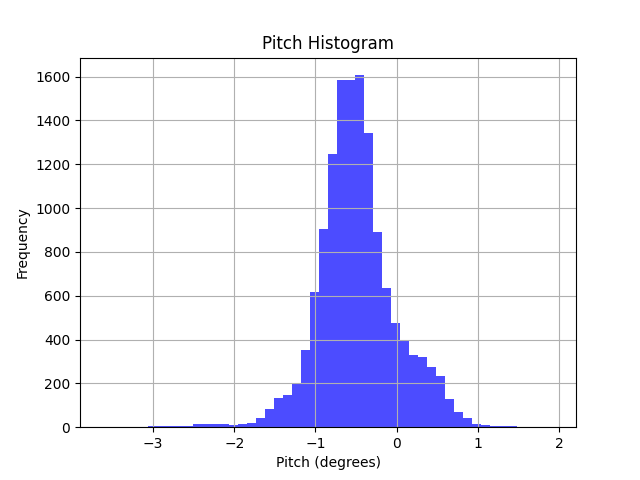}  \\
		
		\raisebox{1.0\normalbaselineskip}[0pt][0pt]{\rotatebox{90}{DrivingStereo \cite{yang2019drivingstereo}}}& 
		\includegraphics[width=\turnwidthnew, trim=0pt 0 40pt 40pt, clip]{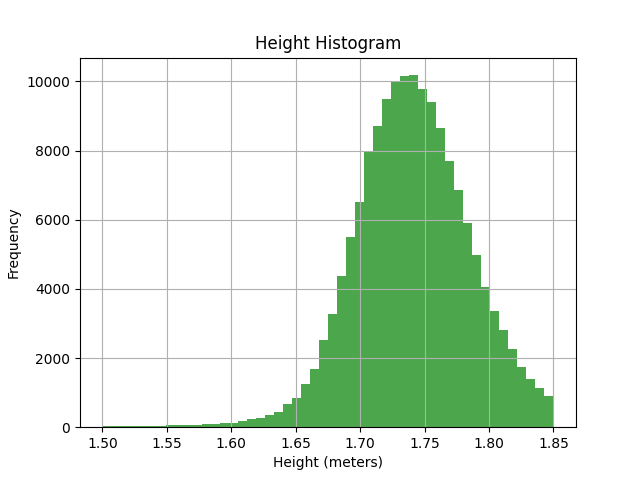} & \includegraphics[width=\turnwidthnew, trim=0pt 0 40pt 40pt, clip]{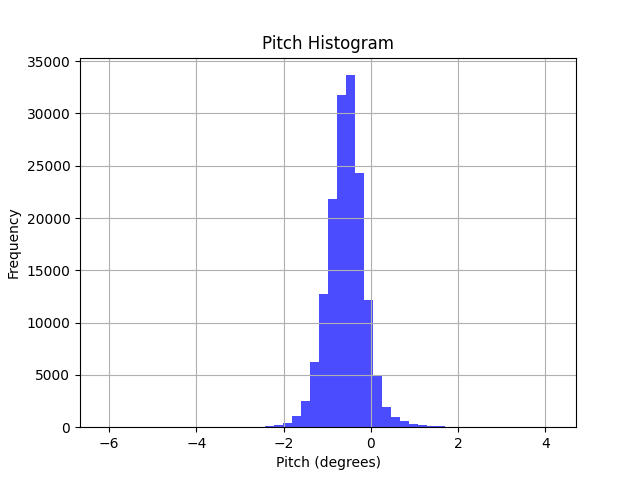}  \\
		
		\raisebox{2.2\normalbaselineskip}[0pt][0pt]{\rotatebox{90}{KITTI \cite{geiger2013vision}}}& 
		\includegraphics[width=\turnwidthnew, trim=0pt 0 40pt 40pt, clip]{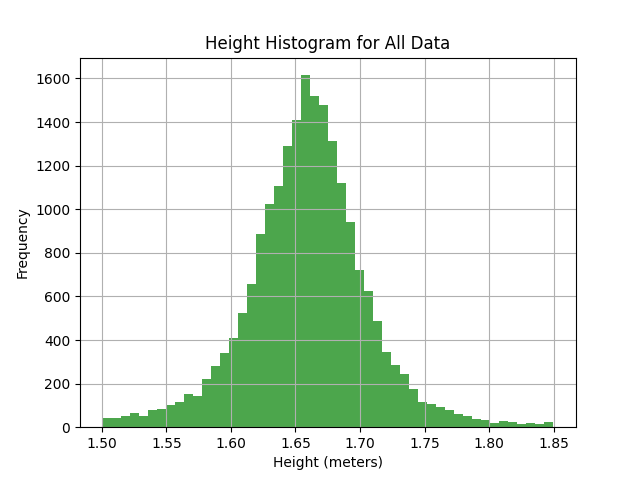} & \includegraphics[width=\turnwidthnew, trim=0pt 0 40pt 40pt, clip]{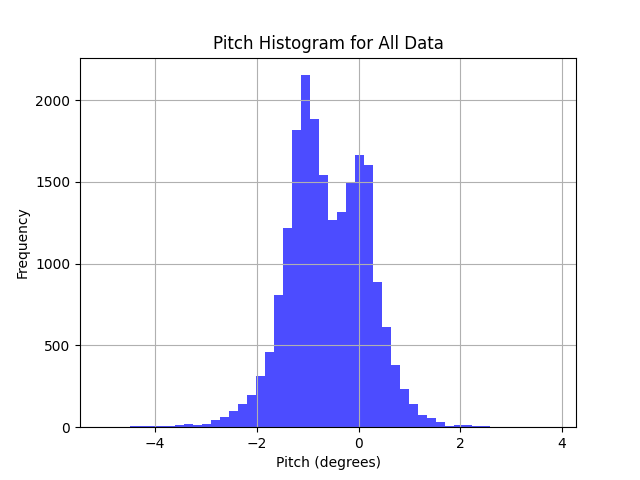}  \\
		
		\raisebox{2.0\normalbaselineskip}[0pt][0pt]{\rotatebox{90}{Waymo \cite{Sun_2020_CVPR}}}& 
		\includegraphics[width=\turnwidthnew, trim=0pt 0 40pt 40pt, clip]{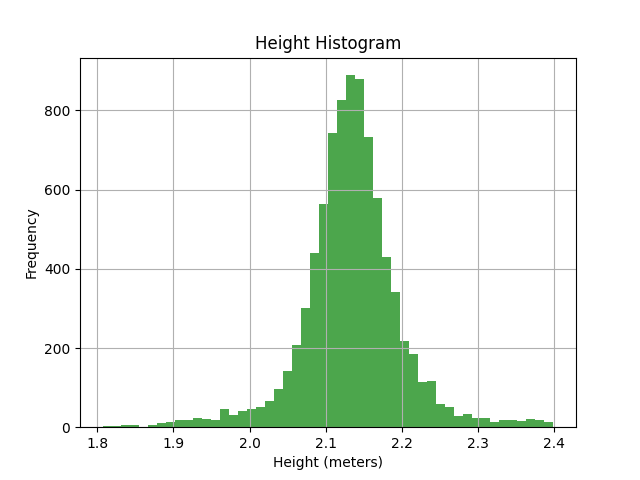} & \includegraphics[width=\turnwidthnew, trim=0pt 0 40pt 40pt, clip]{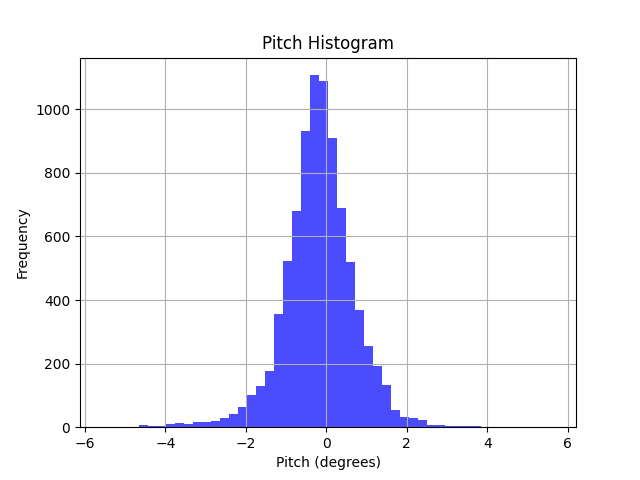}  \\
		
	\end{tabular}
	\caption{\textbf{Calibration histograms.} Height and pitch histograms acquired by calibration procedure described in \cref{alg:calib}. Best viewed zoomned in.}
	\label{fig:calib}
	\vspace{-10pt}
\end{figure}

\noindent{}\textbf{Calibration results.}
Estimated camera height $h$ and camera pitch $\theta$ for each dataset are provided in \cref{tbl:calib}.
Moreover, in \cref{fig:calib} we visualize the histograms obtained with our calibration procedure.
Since DrivingStereo \cite{yang2019drivingstereo} and KITTI \cite{geiger2013vision} do not report extrinsic parameters, we use the estimated values in our proposed canonical transform.
Furthermore, for the DDAD dataset \cite{guizilini20203d}, we observe a significant discrepancy between the official and estimated extrinsics.
Therefore, we use our calibrated parameters, as they align more closely with the ground-truth depth.
In contrast, for Argoverse Stereo \cite{Chang_2019_CVPR} and Waymo \cite{Sun_2020_CVPR}, the estimated values are consistent with the official calibration.

\noindent{}\textbf{KITTI Calibration Issues.}
We would also like to address the problems with KITTI calibration. As shown in \cref{fig:calib}, the pitch histogram for the KITTI dataset does not follow an unimodal distribution.
We believe that this highlights inconsistencies in the official KITTI intrinsics calibration across different recording sequences.
While this observation is not novel \cite{cvivsic2021recalibrating}, it is rarely, if ever, discussed in the context of MDE.
Given that the KITTI dataset remains a cornerstone for MDE evaluation, we believe that this issue should be given increased attention.

\noindent{}\textbf{Evaluation with Per-frame Calibration.}
\begin{table}[t]
	\centering
	\caption{\textbf{Evaluation with per-frame extrinsics calibration.} Vertical -- depth regression via Vertical Canonical Transform - $\mathrm{VCT}_{\bm{\mathcal{C}}}(\cdot)$. Fusion -- depth regression with fusion model. Vertical+ and Fusion+ indicate usage of per-frame camera extrinsic calibration during evaluation. Training datasets on rows, testing datasets on columns. Best results are \textbf{bolded}. In-domain evaluation results are \colorbox{lightgray}{shaded}.}
	\vspace{-5pt}
	\resizebox{\linewidth}{!}{
		\begin{tabular}{cl|ccc|ccc}
			
    		\toprule[0.4mm]
			& \multirow{2}{*}{\textbf{Representation}} & \multicolumn{3}{c|}{\textbf{KITTI}} & \multicolumn{3}{c}{\textbf{DrivingStereo}}\\
			& & $\mathrm{A.Rel}\downarrow$ & \phantom{x}$\mathrm{RMS}\downarrow$ & $\mathrm{\delta}_{1}\uparrow$ & $\mathrm{A.Rel}\downarrow$ & \phantom{x}$\mathrm{RMS}\downarrow$ & $\mathrm{\delta}_{1}\uparrow$\\
			\midrule
			\multirow{4}{*}{\rotatebox{90}{\textbf{KITTI}}} & Vertical & \cellcolor{lightgray}5.70&\cellcolor{lightgray}\textbf{2.58} &\cellcolor{lightgray}95.5 &10.43 &\textbf{5.42} &87.3\\
			& Vertical+ &\cellcolor{lightgray}5.72 &\cellcolor{lightgray}2.62 &9\cellcolor{lightgray}5.6 &10.45 &5.46 &87.1\\
			& Fusion &\cellcolor{lightgray}5.67 &\cellcolor{lightgray}2.61 &\cellcolor{lightgray}95.7 &10.24 &5.66 &87.4\\
			& Fusion+ & \cellcolor{lightgray}\textbf{5.61} & \cellcolor{lightgray}2.60 &\cellcolor{lightgray}\textbf{95.8} & \textbf{10.22} & 5.60 & \textbf{87.5} \\
			\midrule
			\multirow{4}{*}{\rotatebox{90}{\textbf{DStereo}}} & Vertical & 7.52&3.33 &92.6 &\cellcolor{lightgray}3.07 &\cellcolor{lightgray}1.75 &\cellcolor{lightgray}99.5\\
			& Vertical+ &7.31 & 3.28 &92.7 &\cellcolor{lightgray}\textbf{2.78} &\cellcolor{lightgray}\textbf{1.61} &\cellcolor{lightgray}\textbf{99.6}\\
			& Fusion &6.96 &3.17 &92.7 &\cellcolor{lightgray}3.01 &\cellcolor{lightgray}1.76 &\cellcolor{lightgray}99.5\\
			& Fusion+ & \textbf{6.85} & \textbf{3.15} & \textbf{92.7} &\cellcolor{lightgray}2.91 &\cellcolor{lightgray}1.74 &\cellcolor{lightgray}99.6 \\
			\bottomrule[0.4mm]
			
	\end{tabular}}

	\label{tbl:perframe}

\end{table}
Our per-frame calibration procedure offers potentially higher accuracy than median-filtered results.
This is because it better captures ground plane perturbations and vehicle dynamics, which cause slight variations in camera pitch. In \cref{tbl:perframe}, we present evaluation results using per-frame calibration.
While this approach leads to slight performance improvement, the gain is marginal, possibly due to the noise in our calibration process.
Although such additional information about vehicle dynamics and perturbation could theoretically be integrated into a real system using localization with visual odometry or sensor fusion, we exclude these results from the main paper since competing methods do not leverage this information.

\begin{figure}
	\includegraphics[width=\linewidth, trim=40pt 0pt 45pt 0pt, clip]{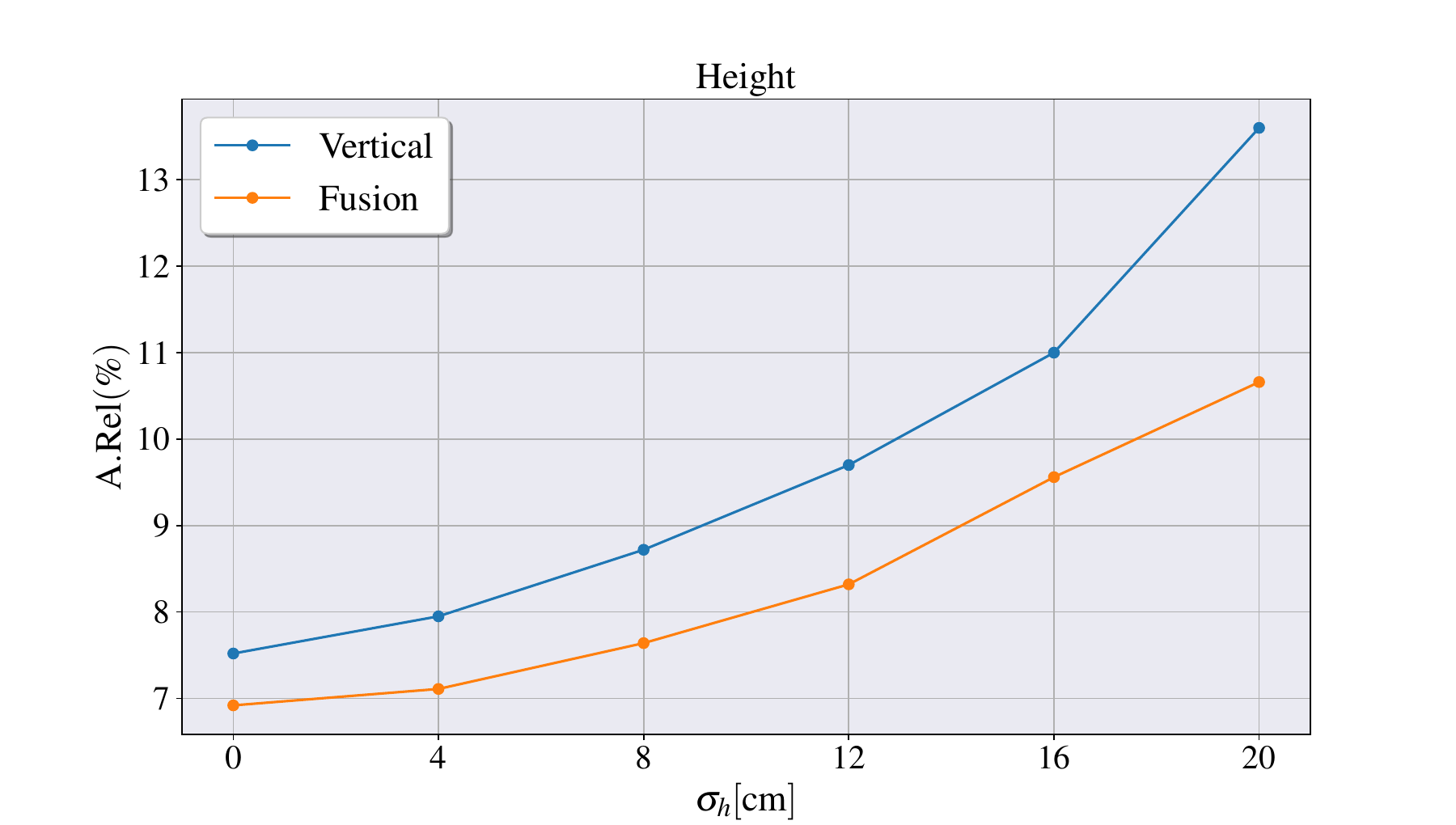}
	\includegraphics[width=\linewidth, trim=40pt 0pt 45pt 0pt, clip]{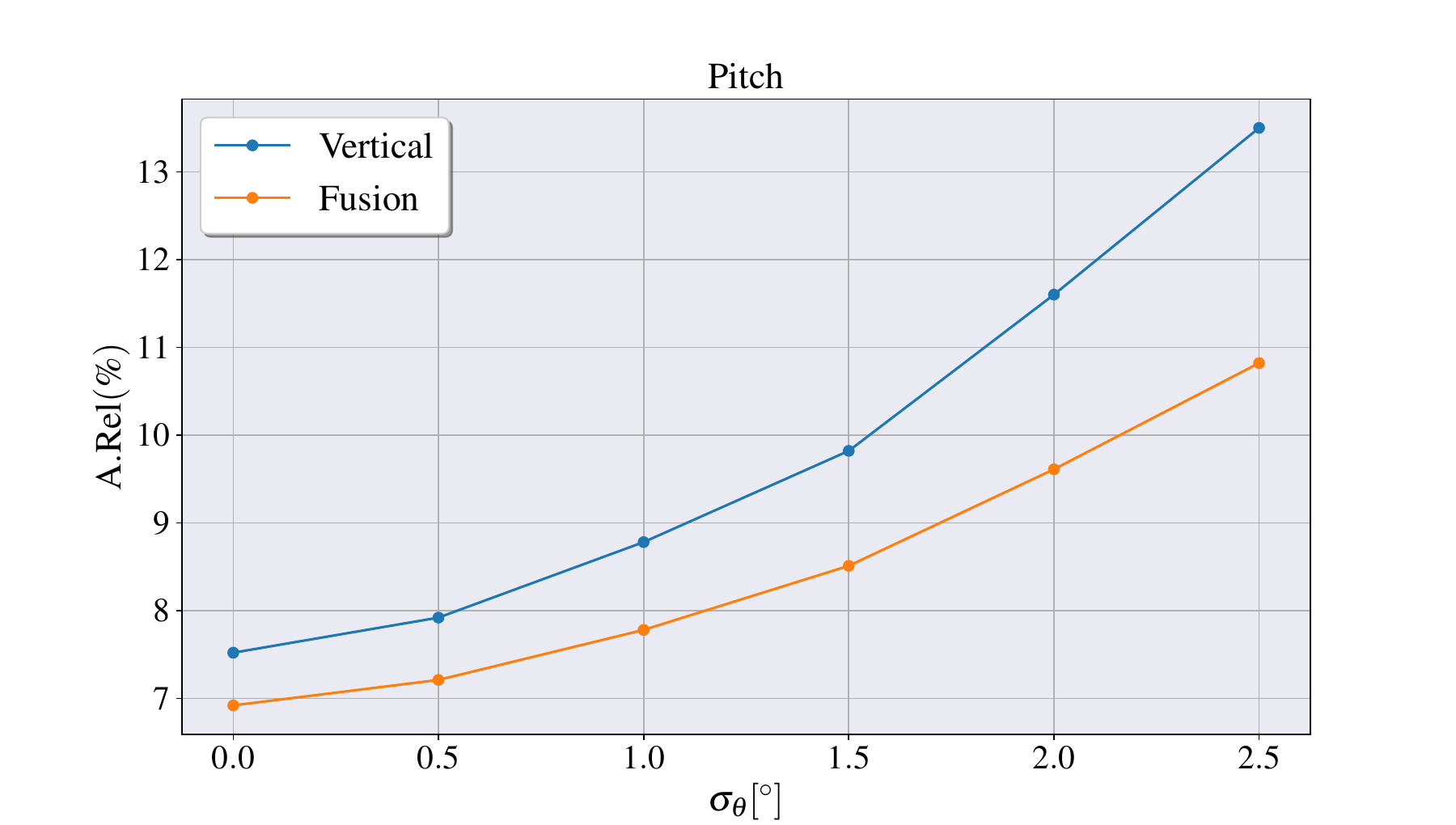}
	\caption{\textbf{Calibration sensitivity.} Sensitivity to Gaussian noise perturbations in extrinsic parameters evaluated on KITTI dataset. Vertical -- depth regression via Vertical Canonical Transform - $\mathrm{VCT}_{\bm{\mathcal{C}}}(\cdot)$. Fusion -- depth regression with fusion model. Train/test dataset combination for both models is DrivingStereo $\rightarrow$ KITTI.}
	\label{fig:calibration_sensitivity}
\end{figure}
\noindent{}\textbf{Calibration Sensitivity.}
Models involving canonical mapping via $\mathrm{VCT}_{\bm{\mathcal{C}}}(\cdot)$ are inherently sensitive to inaccuracies in camera extrinsic parameters.
In \cref{fig:calibration_sensitivity}, we evaluate the $\mathrm{A.Rel}$ metric for our Vertical and Fusion model configurations under varying levels of Gaussian noise in the extrinsic parameters.

While both models exhibit sensitivity to calibration noise, the Fusion model demonstrates a smaller error increase due to its uncertainty-based fusion with depth from $\mathrm{FCT}_{\bm{\mathcal{C}}}(\cdot)$.
However, the Fusion model’s error still increases under noisy conditions, indicating that its uncertainty prediction does not fully compensate for inaccuracies in camera calibration.
This is expected, as camera calibration directly affects the canonical mapping performed by $\mathrm{VCT}_{\bm{\mathcal{C}}}(\cdot)$ and is not internally estimated by the model.

\section{Additional Considerations}
\noindent{}\textbf{Multi-dataset Training.}
Almost all challenges in multi-dataset training for MDE arise from inconsistent and ambiguous perspective geometries, making our approach inherently scalable due to the invariance introduced in canonical spaces.
While we lacked the computational resources to scale the training to the level of a MDE foundation model, in Tab. 3 in main text we demonstrated our method's ability to leverage diverse perspective geometries in the training data induced by geometric augmentations, especially compared to the \textit{Baseline}.

\noindent{}\textbf{Flat Ground Assumption.}
Our method implicitly assumes the ideal ground plane within the  $\mathrm{VCT}_{\bm{\mathcal{C}}}(\cdot)$.
Unlike in GEDepth \cite{yang2023gedepth}, which explicitly models the ground slope, we choose to grant the model greater flexibility.
Motivation behind this choice is straightforward; our proposed canonical transformation is designed to assist the model in resolving ambiguities that diverse training data alone cannot, such as the entanglement of depth and camera parameters.
For other adversarial perturbations, like ground plane imperfections, we do not explicitly model them. 
Instead, we allow the model to internally adjust values in canonical space when it detects these perturbations within highly diverse training data.
Moreover, the uncertainty-based fusion provides the model with additional capabilities to adaptively weight cues based on aleatoric uncertainty, enabling the down-weighting of specific cue in highly uncertain regions.

\vspace{10pt}
\noindent{}\textbf{Acknowledgment.} This research has been funded by the H2020 project AIFORS under Grant Agreement No 952275.

{
	\small
	\bibliographystyle{ieeenat_fullname}
	\bibliography{main}
}
\end{document}